\documentclass[10pt,twocolumn,letterpaper]{article}

\usepackage{iccv}
\usepackage{times}
\usepackage[pdftex]{graphicx}
\usepackage{amsmath}
\usepackage{amssymb}
\usepackage{everyshi}
\usepackage{booktabs}
\usepackage{adjustbox}
\usepackage{subcaption}
\usepackage{multirow}
\usepackage{pifont}
\usepackage{makecell}
\usepackage[table]{xcolor}
\definecolor{lightgray}{rgb}{0.87,0.87,0.87}
\usepackage[accsupp]{axessibility}

\usepackage[pagebackref=true,breaklinks=true,letterpaper=true,colorlinks,bookmarks=false]{hyperref}

\iccvfinalcopy 

\def\iccvPaperID{6798} 

\ificcvfinal\fi

\begin{document}

\title{Open-vocabulary Video Question Answering: A New Benchmark for \\ Evaluating the Generalizability of Video Question Answering Models}

\newcommand*\samethanks[1][\value{footnote}]{\footnotemark[#1]}
\author{
Dohwan Ko\hspace{0.4cm}
Ji Soo Lee\hspace{0.4cm}
Miso Choi\hspace{0.4cm} \\
Jaewon Chu\hspace{0.4cm} 
Jihwan Park\hspace{0.4cm}
Hyunwoo J. Kim\thanks{Corresponding author.}\vspace{0.3cm} \\
Department of Computer Science and Engineering, Korea University\hspace{0.4cm} \\
\tt\small \{ikodoh, simplewhite9, miso8070, allonsy07, jseven7071, hyunwoojkim\}@korea.ac.kr }

\newtheorem{thm}{Theorem}
\newtheorem{rem}[thm]{Remark}
\definecolor{blue_title}{HTML}{BFF2FF}
\definecolor{red_title}{HTML}{FFD7D1}

\maketitle
\ificcvfinal\thispagestyle{empty}\fi

\begin{abstract}
    Video Question Answering (VideoQA) is a challenging task that entails complex multi-modal reasoning.
    In contrast to multiple-choice VideoQA which aims to predict the answer given several options, the goal of open-ended VideoQA is to answer questions without restricting candidate answers.
    However, the majority of previous VideoQA models formulate open-ended VideoQA as a classification task to classify the video-question pairs into a fixed answer set, \textit{i.e.}, closed-vocabulary, which contains only frequent answers (\textit{e.g.}, top-1000 answers).
    This leads the model to be biased toward only frequent answers and fail to generalize on out-of-vocabulary answers.
    We hence propose a new benchmark, Open-vocabulary Video Question Answering (OVQA), to measure the generalizability of VideoQA models by considering rare and unseen answers.
    In addition, in order to improve the model's generalization power, we introduce a novel GNN-based soft verbalizer that enhances the prediction on rare and unseen answers by aggregating the information from their similar words.
    For evaluation, we introduce new baselines by modifying the existing (closed-vocabulary) open-ended VideoQA models and improve their performances by further taking into account rare and unseen answers.
    Our ablation studies and qualitative analyses demonstrate that our GNN-based soft verbalizer further improves the model performance, especially on rare and unseen answers.
    We hope that our benchmark OVQA can serve as a guide for evaluating the generalizability of VideoQA models and inspire future research.
    Code is available at \url{https://github.com/mlvlab/OVQA}.
\end{abstract}
\section{Introduction}
\begin{figure}[t] 
    \centering
    \begin{subfigure}[t]{0.34\linewidth}
        \includegraphics[width=1.0\linewidth]{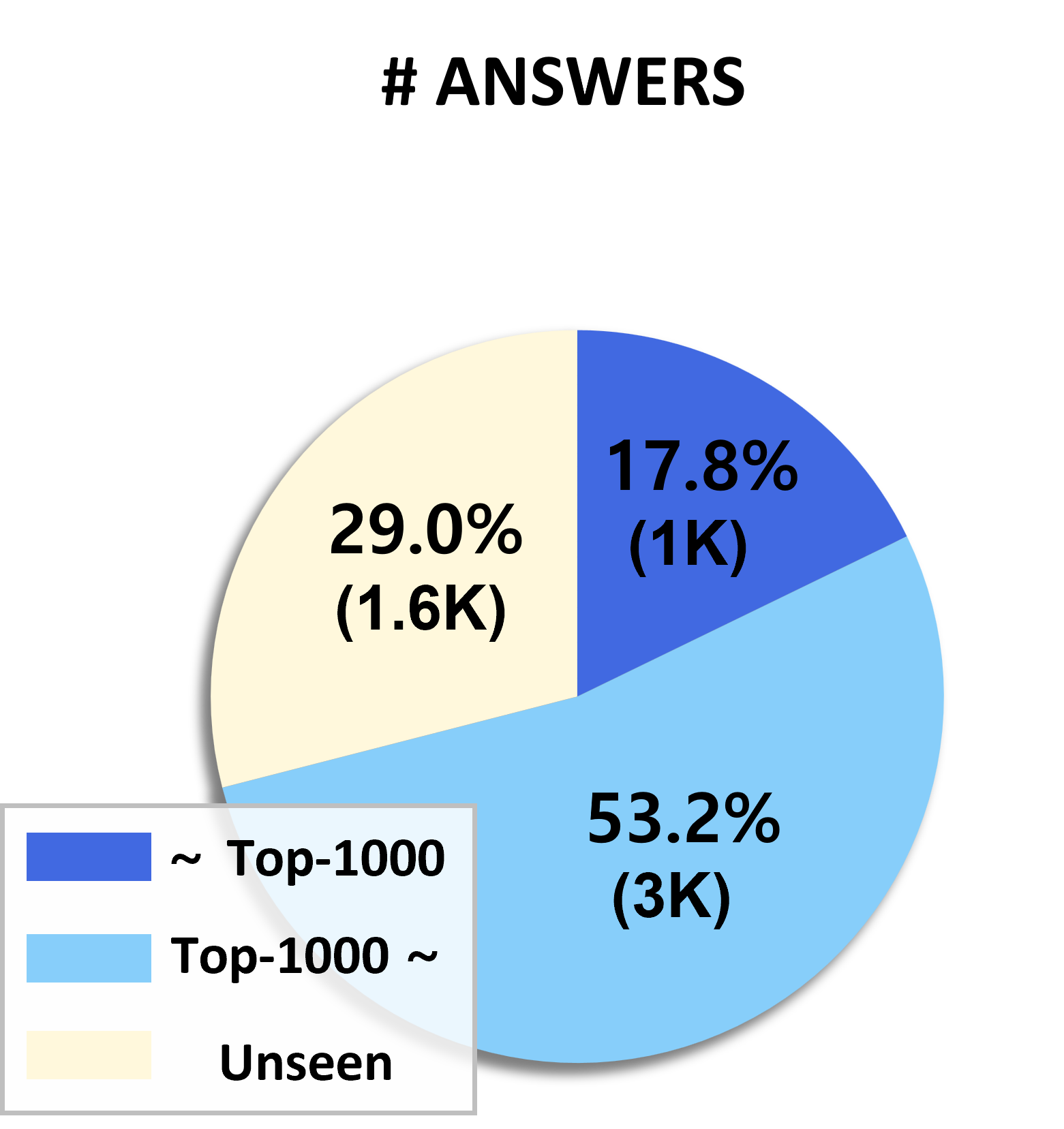}
        \caption{ }
        \label{fig:answers}
    \end{subfigure}
    \begin{subfigure}[t]{0.30\linewidth}
        \includegraphics[width=1.0\linewidth]{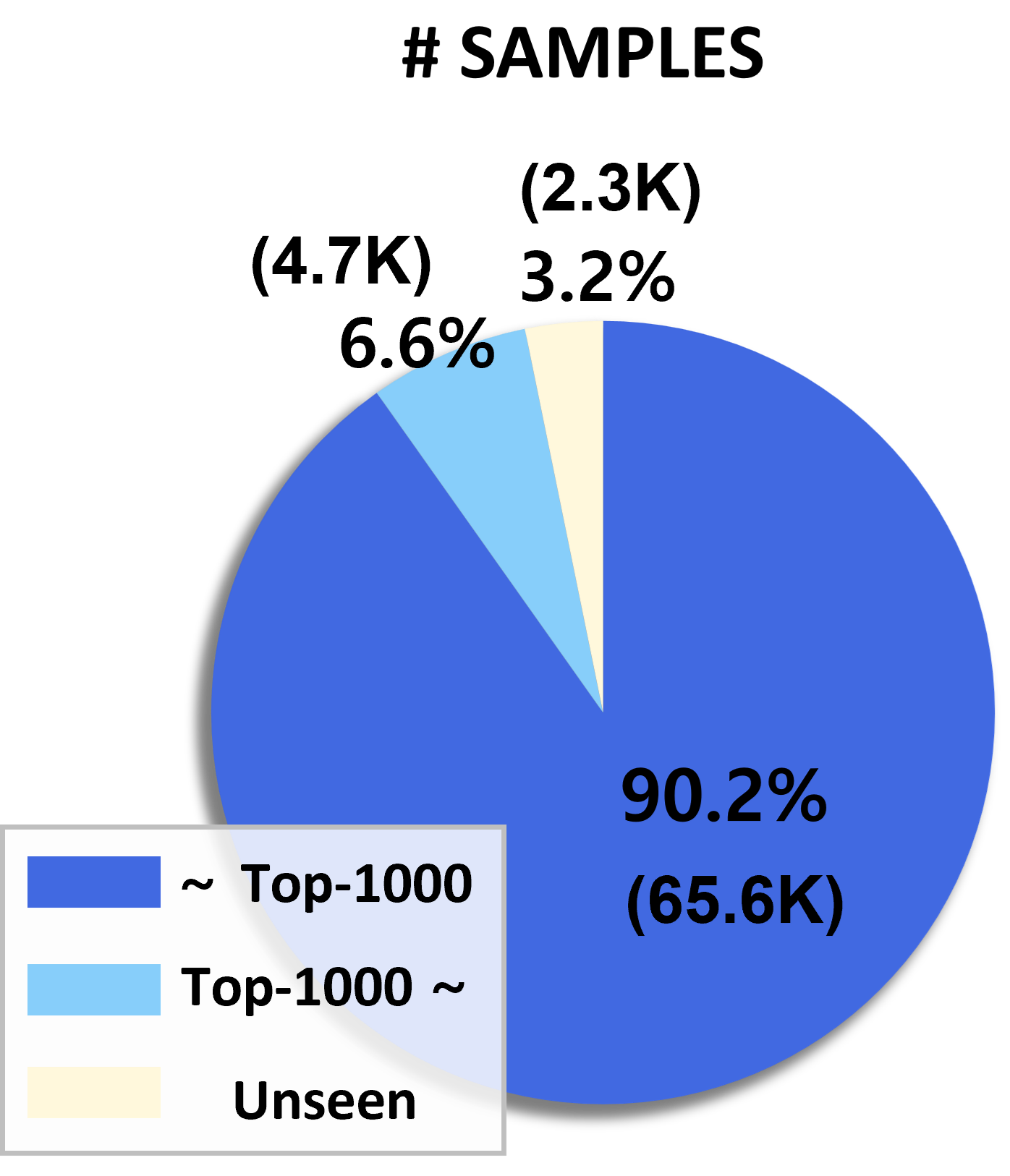}
        \caption{ }
        \label{fig:samples}
    \end{subfigure}
    \begin{subfigure}[t]{0.34\linewidth}
        \includegraphics[width=1.0\linewidth]{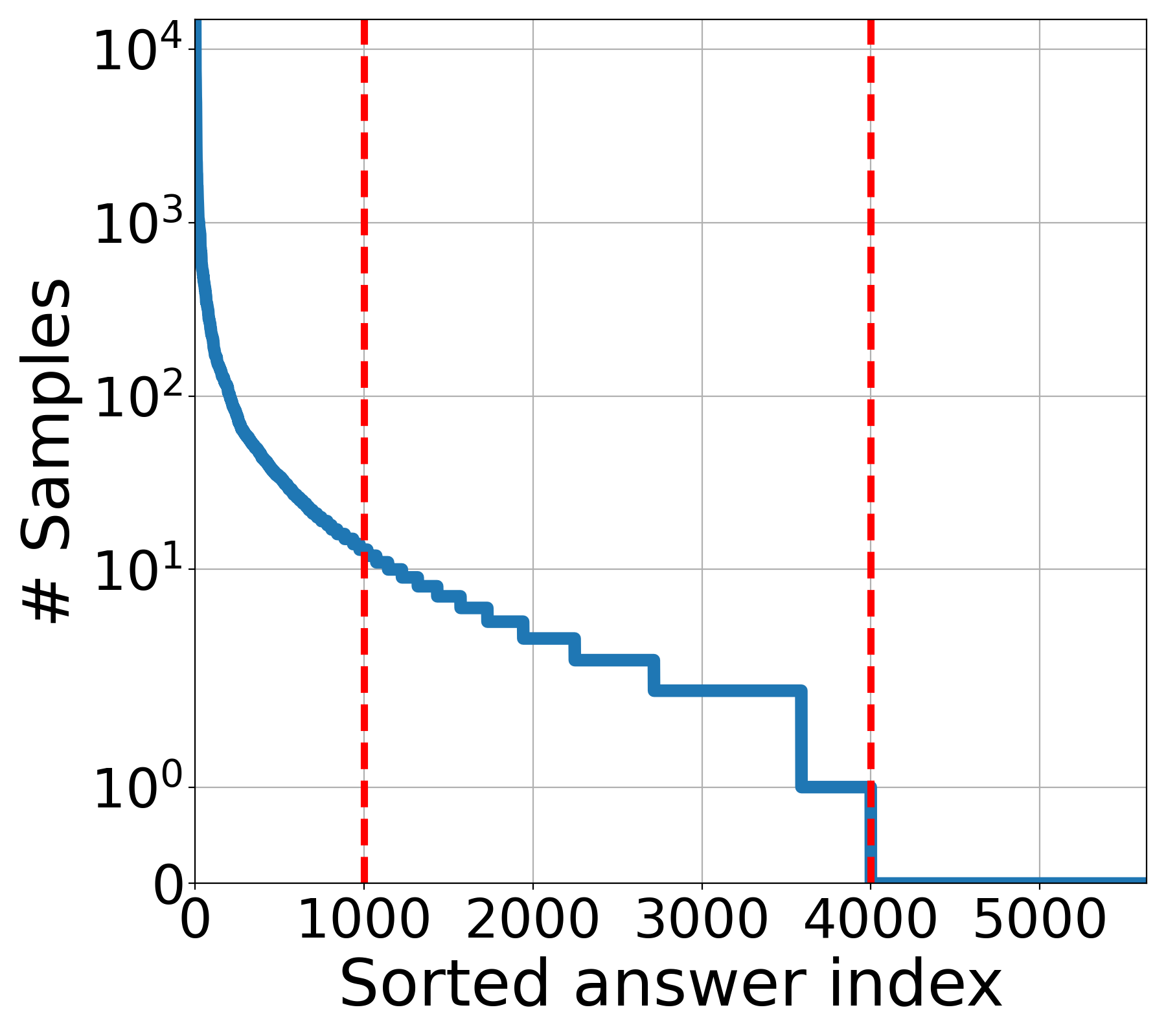}
        \caption{ }
        \label{fig:line}
    \end{subfigure}
    \caption{\textbf{MSRVTT-QA statistics of three answer groups.} 
    Illustration of three different answer groups: the 1000 most frequent answers in the training set ($\sim$ Top-1000), 
    the remaining answers in the training set (Top-1000 $\sim$), and unseen answers which do not exist during training but appear in the test set (Unseen).
    (a) shows the proportion of the number of unique answers in each group.
    (b) shows the proportion of the number of samples in each group.
    (c) shows the distribution of the number of samples for each sorted answer.
    Note that the red lines distinguish each group and the y-axis is an exponential scale.
    }
    \label{fig:msrvtt}
\end{figure}

Video question answering (VideoQA) is a multi-modal understanding task that requires complex reasoning between two modalities to find the correct answer given a video-question pair.
There are usually two task types in VideoQA, multiple-choice and open-ended.
The multiple-choice VideoQA requires the model to select the correct answer among several options.
On the other hand, in open-ended VideoQA, the model needs to predict the answer without restricting candidate vocabulary.

However, most existing VideoQA models~\cite{lei2021less,wang2022all,li2020hero,fu2021violet,zellers2021merlot,le2020hierarchical,yang2022zero} formulate the open-ended VideoQA task as a classification problem with a fixed set of answer candidates which frequently appear in the training set, \textit{e.g.}, top-1000.
Therefore, the out-of-vocabulary answers, not used during training, are automatically regarded as incorrect without any thorough consideration during evaluation.
Fig.~\ref{fig:answers} highlights that the top-1000 answer categories cover about 17.8\% of the answer candidates while they possess about 90.2\% of the total samples overwhelming those of other answer categories in Fig.~\ref{fig:samples}.
This suggests that previous models may show seemingly good performance only with \textit{top-k} answer candidates, yet they, in fact, fail to generalize to rare and unseen answers by ignoring the underrepresented out-of-vocabulary answers.
Such problems have been overlooked since these models have been evaluated in terms of overall performance only.
In other words, the conventional benchmark of open-ended VideoQA does not measure the generalizability and thus leads the model to neglect the realistic setting of class imbalance and unseen answers.
Therefore, a comprehensive benchmark that handles long-tail distribution with unseen answers is necessary.

A long-tail distribution with rare and unseen answers requires few-shot and zero-shot generalization.
Recently, prompt-tuning~\cite{radford2021learning, zhou2022learning, chen2022prompt, jia2022visual, zhou2022conditional} with large-scale pretrained models has drawn attention due to its significant performance gain on zero-shot and few-shot learning.
A line of work~\cite{schick2021exploiting,liu2021gpt,gao2021making,cui2022prototypical,shin2020autoprompt,schick2020automatically,holtzman2021surface,hu2022knowledgeable} enables fine-tuning the model in a parameter-efficient manner by retaining the Masked Language Modeling (MLM) objective leveraged in the pretraining phase.
In other words, the model is asked to fill in [MASK] tokens for its downstream objectives.
Subsequently, the concept of \textit{verbalizer} was introduced by \cite{schick2021exploiting} to manually bridge the original label and its corresponding words to be filled in [MASK], \textit{e.g.}, filling the word `great' in [MASK] to predict the label POSITIVE in sentiment classification.
To reduce the human labor, search-based verbalizers~\cite{gao2021making,schick2020automatically,shin2020autoprompt} have been proposed. 
Current works~\cite{cui2022prototypical,hambardzumyan2021warp,zhang2021differentiable} adopt soft verbalizers which consist of learnable tokens to find optimal embeddings during training.
However, verbalizers for unseen answers have been less explored in the literature.

To this end, we introduce a new benchmark of open-ended VideoQA, named Open-vocabulary Video Question Answering (OVQA), to define the task under a more real-world setting with rare and unseen answers.
In contrast to previous approaches which focus only on frequent answers, OVQA requires the model to predict rare and out-of-vocabulary answers.
In OVQA, to address the problem of bias towards frequent answers, we propose a novel graph neural network (GNN)-based soft verbalizer to smooth the original answer embeddings by aggregating the information of similar words from an external knowledge base.
Specifically, the GNN-based soft verbalizer learns how to smooth the original answers with their neighborhood words in the training phase and is adapted to the test phase based on the learned smoothing function during training to enhance the prediction for the unseen answers.

In our experiments, on four benchmark open-ended VideoQA datasets (MSVD-QA, ActivityNet-QA, TGIF-QA, and MSRVTT-QA), we develop OVQA baseline models with an additional answer encoder and improve their performances by taking into account rare and unseen answers as well.
Also, our extensive ablation studies demonstrate that GNN-based soft verbalizer is generally adaptable to various backbone models and effectively reduces the bias towards frequent answers.

\noindent To sum up, our \textbf{contributions} are as follows:
\begin{itemize}
    \item[\textbullet] We propose a new benchmark of open-ended VideoQA, OVQA, to evaluate models' generalizability under a long-tail distribution, including unseen answers.
    \item[\textbullet] We also present a novel GNN-based soft verbalizer to smooth answers on the answer graphs augmented with an external knowledge base.
    \item[\textbullet] Our experiments show that baselines are consistently improved by our simple modification with an additional answer encoder to handle out-of-vocabulary answers.
    \item[\textbullet] Extensive ablation studies and qualitative analyses demonstrate that GNN-based soft verbalizer is broadly  applicable and alleviates the bias problem toward frequent answers.
\end{itemize}
\section{Open-vocabulary video question answering}
In this section, we introduce a new benchmark, Open-vocabulary Video Question Answering (OVQA), to tackle the problem of a common practice that formulates open-ended VideoQA as a classification task with fixed answer candidates.

\begin{figure*}[t] 
    \centering
    \begin{subfigure}[t]{0.425\linewidth}
        \includegraphics[width=1.0\linewidth]{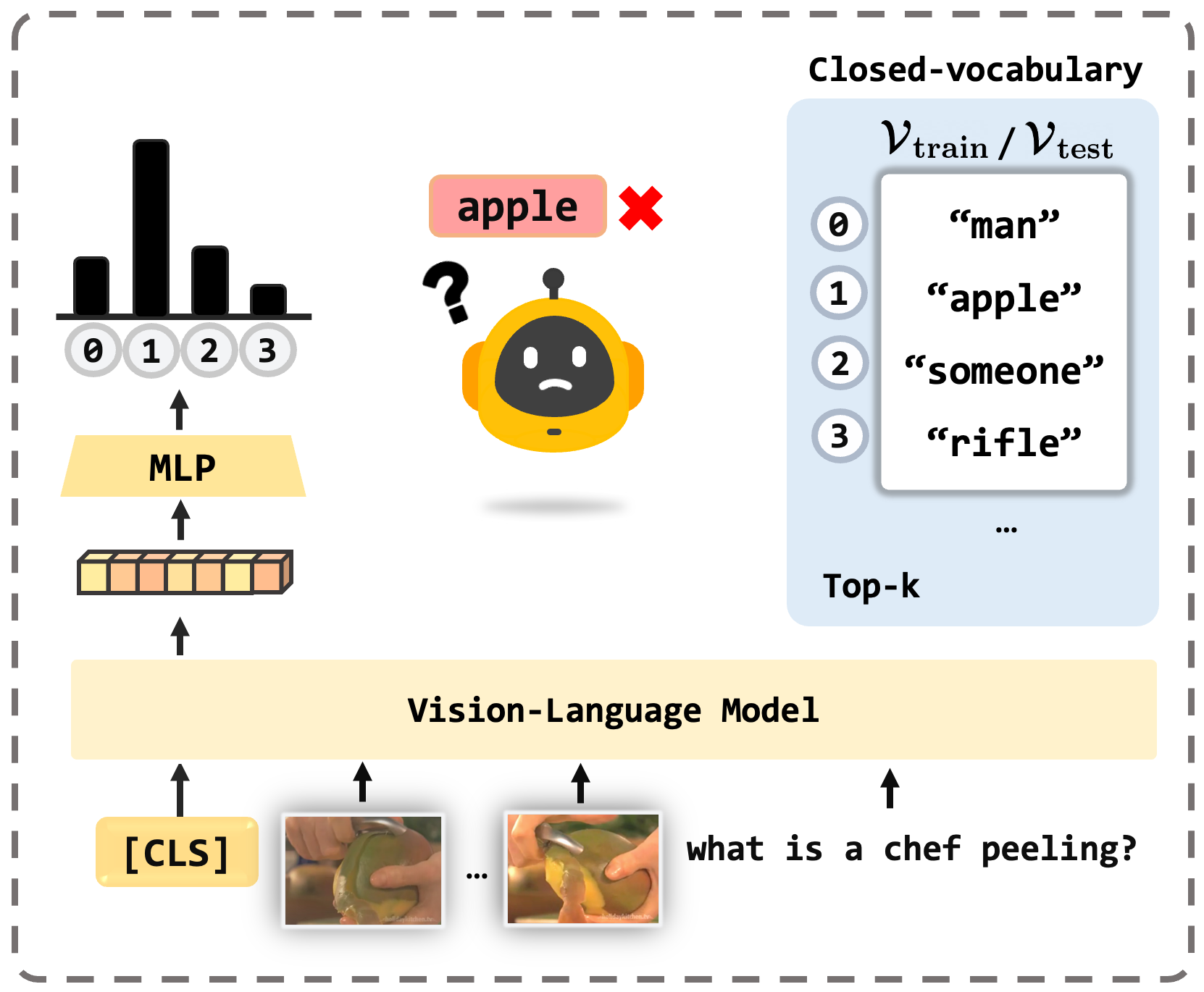}
        \caption{Closed-vocabulary VideoQA (CVQA)}
        \label{fig:cvqa}
    \end{subfigure}
    \begin{subfigure}[t]{0.565\linewidth}
        \includegraphics[width=1.0\linewidth]{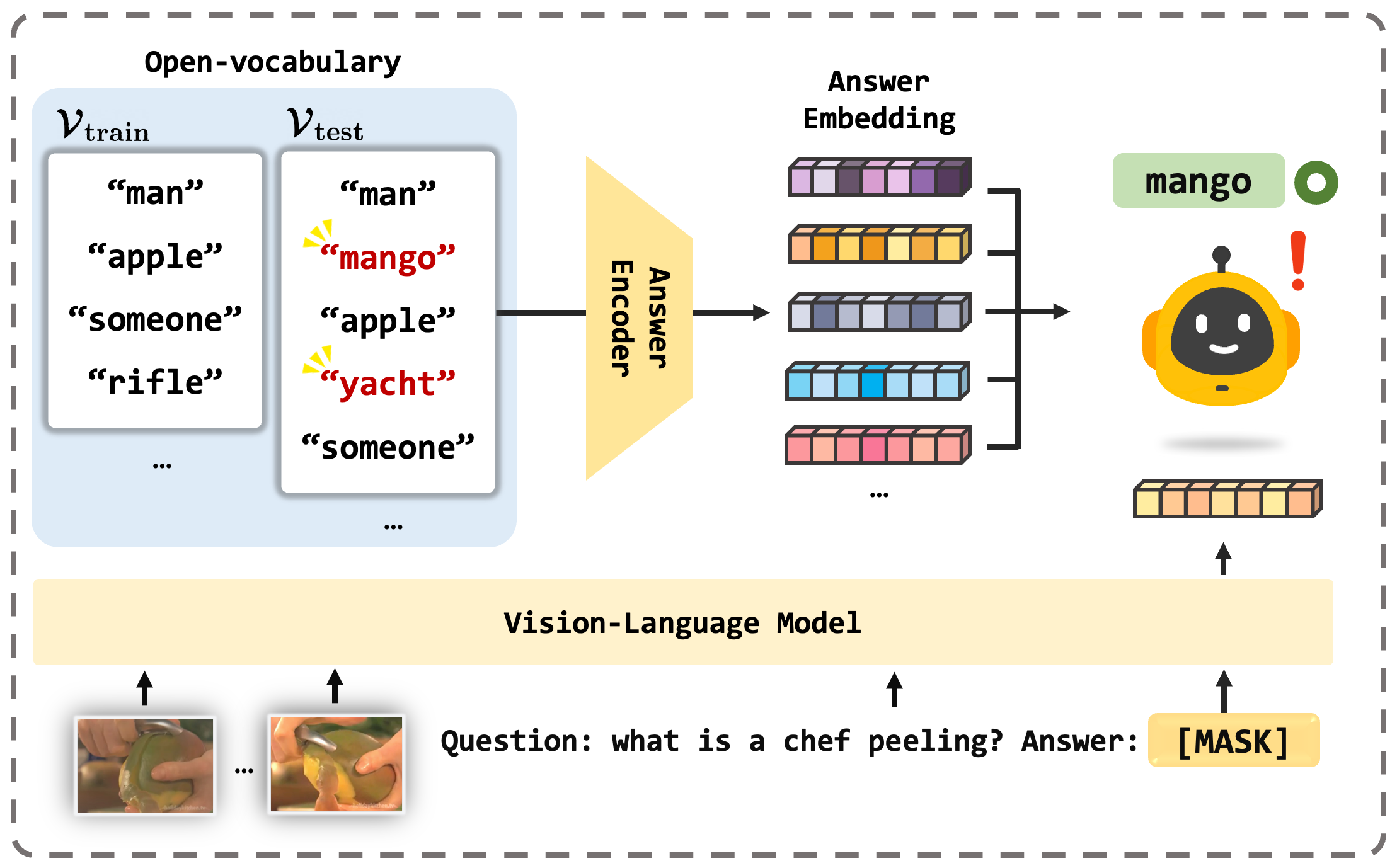}
        \caption{Open-vocabulary VideoQA (OVQA)}
        \label{fig:ovqa}
    \end{subfigure}
    \caption{\textbf{Comparison of CVQA and OVQA.}
    (a) The output feature of [CLS] token is fed to an MLP to calculate the logits over the fixed \textit{top-k} answer candidates (closed-vocabulary) thus it fails to select the out-of-vocabulary answers in the test phase.
    (b) On the other hand, in our OVQA setting, the model chooses the answer based on the similarities between the output feature of [MASK] token and the answer embeddings.
    Therefore, the model can predict the answer although the answer is unseen at the training phase.
    }
    \label{fig:comparison}
\end{figure*}

\subsection{Open-ended VideoQA}
\label{subsec:oeqa}

Unlike multiple-choice VideoQA where a model needs to choose one answer among the given five options, the open-ended VideoQA task aims to predict the answer without any candidate answers.
However, previous works~\cite{lei2021less,wang2022all,li2020hero,fu2021violet,zellers2021merlot,le2020hierarchical,yang2022zero} formulate open-ended VideoQA as a classification problem with a predefined answer set containing fixed candidate answers.
We call this setting Closed-vocabulary Video Question Answering (CVQA) for the rest of our paper.
Usually, in CVQA, they construct an answer vocabulary based on the frequencies of answers in the training set, \textit{e.g.}, top-1000 answers.
As a result, the out-of-vocabulary answers not used for training will be considered incorrect during evaluation.
In other words, previous models learn to predict only the \textit{top-k} answers that frequently appear in the training set and ignore rare or unseen answers.
This leads the model to be biased toward frequent answers and fail to generalize on rare and unseen answers, \textit{i.e.}, they \textit{memorize} the answers rather than \textit{generalize}.

We first categorize all the answers from four benchmark datasets (MSVD-QA, ActivityNet-QA, TGIF-QA, and MSRVTT-QA) based on how many $\langle$\textit{video}, \textit{question}, \textit{answer}$\rangle$ triplets from the training set they appear in: \textit{unseen} (0 times), \textit{rare} (1 $\sim$ 10), \textit{common} (11 $\sim$ 100), and \textit{base} (101 $\sim$).
The \textit{unseen} answers are only present in the test set while the answers of other categories are seen in the training set but may or may not appear in the test set.
Tab.~\ref{tab:statistics} shows the number of unique answers for each category.
For an example of MSRVTT-QA, in CVQA, top-1000 answers only include base and common answers.
Therefore, we propose a new benchmark of open-ended VideoQA to provide an opportunity to consider the rare and even unseen answers.
\begin{table}[t!]
    \centering
    \begin{adjustbox}{width=\linewidth}
    \begin{tabular}{l|c c c c}
        \toprule
        & MSVD-QA & MSRVTT-QA & TGIF-QA & ActivityNet-QA \\
        \midrule
        \midrule 
        Base (101 $\sim$)   & 41 & 205 & 38 & 26 \\
        Common (11 $\sim$ 100) & 333 & 937 & 210 & 275 \\
        Rare (1 $\sim$ 10)   & 1,478 & 2,858 & 1,292 & 1,353 \\
        Unseen (0) & 391 & 1,632 & 206 & 1,378 \\
        \midrule
        Total   & 2,243 & 5,632 & 1,746 & 3,032 \\
        \bottomrule
    \end{tabular}
    \end{adjustbox}
    \caption{\textbf{Answer statistics.}
    We report the number of answers for each category: base, common, rare, and unseen.
    }
    \label{tab:statistics}
\end{table}
\subsection{Task definition}

We here introduce a new benchmark, Open-vocabulary Video Question Answering (OVQA), which considers not only the frequent answers but also the rare or unseen answers.
Prior studies in CVQA have calculated logits with an MLP on video-question multi-modal features for each class label that corresponds to the individual answer candidate as shown in Fig.~\ref{fig:cvqa}.
Nevertheless, they fail to determine the logit scores of the out-of-vocabulary answers that are unseen in the training set.
To consider all the answer vocabularies in OVQA, we also introduce new baselines which further encode the answer features and calculate the similarity between the video-question features and the encoded answer features. 
This enables the open-vocabulary setting which is capable of handling unseen answers as illustrated in Fig.~\ref{fig:ovqa}.
As a result, unlike previous CVQA models memorizing only frequent answers, the goal of OVQA is to consider all the open-vocabulary answers and evaluate the model performance and its generalizability without ignoring rare or unseen answers.

Similar to the CVQA evaluation metric, we use the accuracy (\%) metric for OVQA.
Yet, we report the total accuracy as well as the accuracy for each answer category (base, common, rare, and unseen).
We also introduce a mean accuracy (mAcc), averaging the accuracy for each unique answer, to assess the generalizability of the model.
\subsection{Comparison with other benchmarks}

There have been several attempts to evaluate the visual question answering models under out-of-distribution (OOD) settings since a number of studies have revealed that most existing models rely extremely on dataset bias to answer questions~\cite{agrawal2018don,kervadec2021roses,niu2021counterfactual, ramakrishnan2018overcoming, cadene2019rubi}.
For example, in Visual Question Answering, \cite{agrawal2018don} proposed VQA-CP v2, a new split of VQA v2~\cite{goyal2017making}, by changing the answer distribution for each question type between train and test splits, and pointed out that previous models are vulnerable to such distribution shifts.
Also, GQA-OOD~\cite{kervadec2021roses} re-organized GQA dataset~\cite{hudson2019gqa} and introduced a new benchmark with more comprehensive evaluation metrics (\textit{e.g.}, acc-tail and acc-head).
However, these benchmarks did not investigate the \textit{unseen} answers, which cannot assess the models’ zero-shot adaptability.
In Video Question Answering, NExT-QA~\cite{xiao2021next} introduced open-form video question answering which requires the model to generate the answer, \textit{i.e.}, a generation problem, without fixed answer candidates.

In contrast to previous efforts, our OVQA aims to assess the models' generalizability under a long-tail distribution including out-of-vocabulary answers, \textit{i.e.}, few-shot and zero-shot adaptability.
The term `open-vocabulary' means that a model is required to predict answers that are \textit{unseen} during training by comparing the similarity between the video-question feature and the answer feature.
With a sufficiently large number of unseen vocabulary, we define Open-vocabulary VideoQA.
\section{GNN-based soft verbalizer}

\begin{figure}[t] 
    \centering
    \includegraphics[width=\linewidth]{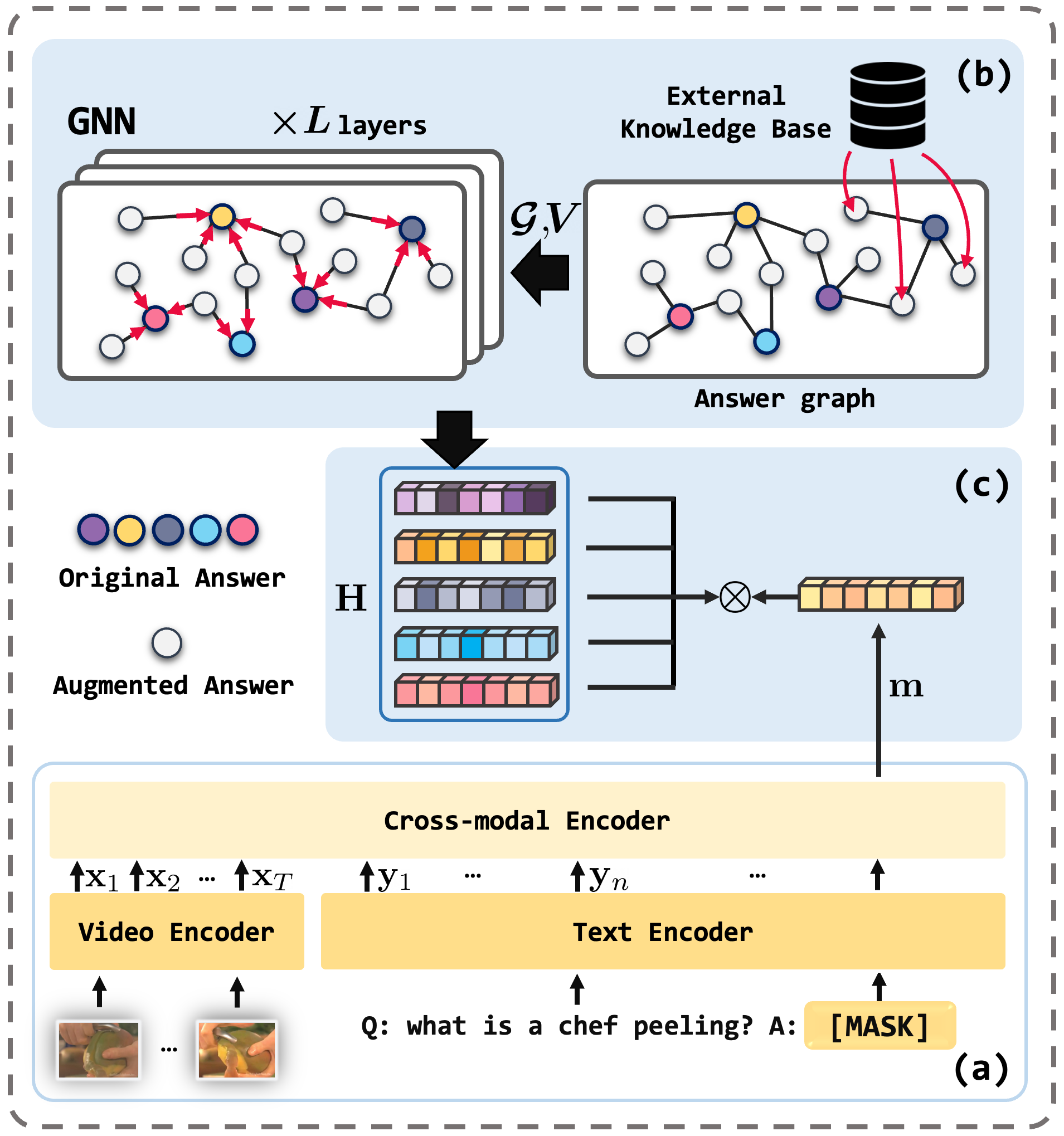}
    \caption{\textbf{Overall architecture.} 
    \textbf{(a) Video-question encoding:} a video-question pair is first encoded through a backbone architecture and the output feature of [MASK] token, $\mathbf{m} \in \mathbb{R}^D$, is extracted.
    \textbf{(b) GNN-based soft verbalizer:} an answer graph is constructed with both original answers and their augmented words from an external knowledge base, and GNN aggregates their information.
    \textbf{(c) Similarity calculation:} we finally calculate the similarity (denoted as $\otimes$) between smoothed answer embeddings $\mathbf{H}_\text{train}$ (or $\mathbf{H}_\text{test}$) and [MASK] token output feature $\mathbf{m}$.
    }
    \label{fig:main}
\end{figure}

By adopting an additional answer encoder to extract answer embeddings to enable OVQA, it is worth designing a way to fine-tune the answer embeddings.
To achieve this, we propose a novel GNN-based soft verbalizer.
The goal of our framework is learning to smooth the original answer candidates with their similar words augmented by an external knowledge base (\textit{e.g.}, GloVe~\cite{pennington2014glove} and ConceptNet~\cite{speer2017conceptnet}).
Thus it helps the model enhance the prediction of rare or unseen answers and improves its generalizability by aggregating information from their neighborhoods.
The overall architecture is illustrated in Fig.~\ref{fig:main}.
We first briefly summarize the basic concepts of the verbalizer and GNNs, and then delineate our framework.

\subsection{Preliminaries}
\label{subsec:preliminaries}

\noindent \textbf{Verbalizer.}
Large-scale foundation models like BERT~\cite{devlin2018bert}, CLIP~\cite{radford2021learning}, and GPT~\cite{brown2020language} have shown remarkable performance on various domains and tasks, and thus ways to fine-tune those effectively and efficiently have also gained attention.
For example, when fine-tuning on sentiment classification, a common practice is to predict the label (POSITIVE or NEGATIVE) with a task-specific classification head (usually MLP) on [CLS] token of a given sentence.
Nonetheless, this scheme does not fully leverage the pretraining objective, \textit{i.e.}, MLM, and its pretrained layer. 
It discards the MLM head and newly adopts the classification head, which would be trained from scratch with a classification loss, on top of [CLS] token.

To effectively utilize the pretrained MLM head, \cite{schick2021exploiting} reformulated an input sentence into a \textit{cloze} form and implemented prediction by filling in the [MASK] token.
In this literature, the mapping from the label space (POSITIVE or NEGATIVE) to the vocabulary (`great' or `terrible') to be filled into the [MASK] token is called the \textit{verbalizer}.
Recent studies~\cite{hu2022knowledgeable,wang2022automatic} about the verbalizer have proposed one-to-many mapping with similar words from the external knowledge base, \textit{e.g.}, (POSITIVE $\rightarrow$ `great', `perfect', `fun', and `brilliant') and (NEGATIVE $\rightarrow$ `terrible', `awful', `disappointing', and `not').
Also, to deal with the limitations of such hard verbalizers that use discrete label words, \cite{cui2022prototypical, hambardzumyan2021warp, zhang2021differentiable} introduced soft verbalizers by adopting learnable label embeddings.

\noindent \textit{Remarks.}
Unlike prompt-tuning which maps the word to embedding by appending several learnable tokens at the input-level, the soft verbalizer maps the word feature to word feature in the embedding space, while the hard verbalizer maps the word to word in the word-level.

\noindent \textbf{Graph Neural Networks (GNNs).}
A graph is denoted as $\mathcal{G} = (\mathcal{V}, \mathcal{E})$, where $\mathcal{V}$ is a set of nodes and $\mathcal{E}$ is a set of edges.
Each node $i \in \mathcal{V}$ has a node feature vector $v_i \in \mathbb{R}^D$.
A set of neighborhoods of the $i$-th node including itself is defined as $\mathcal{N}_i = \{i\} \cup \{j \in \mathcal{V}|(i, j) \in \mathcal{E}\}$.
The majority of current GNNs~\cite{gilmer2017neural,kipf2016semi} use message-passing frameworks to train graph-structured data as:
\begin{equation}
    \mathbf{h}_i^{(l)} = \sigma \left(\mathbf{W}^{(l)} \cdot \text{AGGREGATE}\left(\mathbf{h}_j^{(l-1)}: j \in \mathcal{N}_i\right) \right),
    \label{eq:message}
\end{equation}
where $\mathbf{h}_i^{(l)}$ is a hidden representation of the $i$-th node on the $l$-th layer, $\mathbf{h}_i^{(0)}$ is an input feature of the $i$-th node, and $\mathbf{W}^{(l)}$ is a learnable weight matrix on the $l$-th layer.
AGGREGATE is an aggregation function defined differently by the model, and $\sigma$ is a non-linear activation function.
$L$-layer GNN is conducted by propagating the input features through Eq.~\eqref{eq:message} $L$ times.

Latest studies~\cite{li2018deeper,wu2019simplifying} have shown that most existing GNNs such as GCN~\cite{kipf2016semi} and GAT~\cite{velivckovic2017graph} effectively learn to propagate information and capture meaningful patterns in the graph when the connected nodes have similar characteristics.
We hence adopt GNN to learn how to smooth the original answer with its similar words and apply it to the test vocabulary answers to adequately handle the rare or unseen answers by smoothing them with their neighborhoods.
\subsection{Overall architecture}

Our model is based on FrozenBiLM~\cite{yang2022zero} consisting of three components: a video encoder, a text encoder, and a cross-modal encoder.

\noindent \textbf{Video encoder.}
Each input video is divided into $T$ frames and each frame is fed into CLIP ViT-L/14~\cite{radford2021learning, dosovitskiy2020image} to extract the features denoted as $\mathbf{X} = \{\mathbf{x}_t\}_{t=1}^T \in \mathbb{R}^{T \times D}$, where $D$ is a feature dimension.

\noindent \textbf{Input prompt and text tokenizer.}
The input text prompt for OVQA is formulated as a \textit{cloze} form~\cite{schick2021exploiting, taylor1953cloze}, \textit{i.e.}, the model is expected to fill in a mask token in the input prompt. 
[CLS] and [SEP] tokens are inserted at the beginning and the end of each sequence.
Textual subtitles attained from automatic speech recognition (ASR) can be optionally appended.
The prompt is as follows: {\small \textbf{``\texttt{[CLS] Question: <Question>? Answer: [MASK]. Subtitles: <Subtitles> [SEP]}''}}.
Each prompt sequence is tokenized to $\mathbf{Y} = \{\mathbf{y}_n\}_{n=1}^{N} \in \mathbb{R}^{N \times D}$ by DeBERTa~\cite{he2020deberta} tokenizer, where $N$ is the number of tokens.

\noindent \textbf{Cross-modal encoder.}
The visual feature $\mathbf{X}$ and text feature $\mathbf{Y}$ are forwarded to the cross-modal encoder.
The model is optimized by the masked language modeling (MLM) objective and we especially denote the output feature of [MASK] token as $\mathbf{m} \in \mathbb{R}^D$.
Then, our model compares the similarity between $\mathbf{m}$ and the answer features also encoded by DeBERTa tokenizer.
Fig.~\ref{fig:main} illustrates our overall architecture.

In contrast to CVQA whose train and test vocabulary sets are consistent with each other (\textit{top-k} frequent answers), we consider two different vocabulary sets $\mathcal{V}_\text{train}$ and $\mathcal{V}_\text{test}$ respectively where the former covers the entire answers from the training set and the latter contains the answers even unseen at the training phase.
We further develop several OVQA baselines by modifying a classfication head.
In details, instead of using MLP as the classification head, we replace it with the similarity calculation between video-question multi-modal features and answer embeddings.
\subsection{Answer graph construction}

We first construct an \textit{answer graph} from an external knowledge base to be used for a GNN-based soft verbalizer.
We denote a neighborhood construction function of the original answer $a$ as $n(a)$.
Note that $n(a)$ may be considered as an one-to-many mapping verbalizer introduced in Sec.~\ref{subsec:preliminaries}.
$n(a)$ is composed of the nearest neighborhood words of $a$ from GloVe~\cite{pennington2014glove}.
Then, we augment them into one node set as:
\begin{equation}
    \begin{split}
        & \mathcal{V}_\text{train}^{(k)} = \{j|j \in n(i) \:\:\text{and}\:\: i \in \mathcal{V}_\text{train}^{(k-1)}\} \cup \mathcal{V}_\text{train}^{(k-1)} \\
        & \mathcal{V}_\text{test}^{(k)} = \{j|j \in n(i) \:\:\text{and}\:\: i \in \mathcal{V}_\text{test}^{(k-1)}\} \cup \mathcal{V}_\text{test}^{(k-1)},
    \end{split}
\end{equation}
where $\mathcal{V}_\text{train}^{(0)} = \mathcal{V}_\text{train}$ and $\mathcal{V}_\text{test}^{(0)} = \mathcal{V}_\text{test}$, \textit{i.e.}, original train and test vocabulary sets.
Also, the set of edges is defined as:
\begin{equation}
    \begin{split}
        & \mathcal{E}_\text{train}^{(k)} = \{(j, i)|j \in n(i) \:\:\text{and}\:\: i \in \mathcal{V}_\text{train}^{(k-1)}\} \\
        & \mathcal{E}_\text{test}^{(k)} = \{(j, i)|j \in n(i) \:\:\text{and}\:\: i \in \mathcal{V}_\text{test}^{(k-1)}\}.
    \end{split}
\end{equation}
Then, the answer graph is as follows:
\begin{equation}
    \mathcal{G}_\text{train}^{(K)} = (\mathcal{V}_\text{train}^{(K)}, \mathcal{E}_\text{train}^{(K)}), \:\:\:
    \mathcal{G}_\text{test}^{(K)} = (\mathcal{V}_\text{test}^{(K)}, \mathcal{E}_\text{test}^{(K)}).
\end{equation}
Note that $\mathcal{G}_\text{train}^{(K)}$ and $\mathcal{G}_\text{test}^{(K)}$ take into account $K$-hop neighborhoods for each answer, and we use $K = 2$ to consider up to 2-hop neighborhoods.
Also, the edges directly connected in-between the original answers are dropped.
\begin{table*}[t!]
    \centering
    \setlength{\tabcolsep}{3.5pt}
    \begin{adjustbox}{width=\linewidth}
    \begin{tabular}{c|c c c c c c | c c c c c c | c c c c c c | c c c c c c }
        \toprule
        \textbf{Models} & \multicolumn{6}{c|}{\textbf{MSVD-QA}} & \multicolumn{6}{c|}{\textbf{ActivityNet-QA}} & \multicolumn{6}{c|}{\textbf{TGIF-QA}} & \multicolumn{6}{c}{\textbf{MSRVTT-QA}} \\
        & \textbf{B} & \textbf{C} & \textbf{R} & \textbf{U} & \textbf{T} & \cellcolor[HTML]{C0C0C0}{\textbf{M}} & \textbf{B} & \textbf{C} & \textbf{R} & \textbf{U} & \textbf{T} & \cellcolor[HTML]{C0C0C0}{\textbf{M}} & \textbf{B} & \textbf{C} & \textbf{R} & \textbf{U} & \textbf{T} & \cellcolor[HTML]{C0C0C0}{\textbf{M}} & \textbf{B} & \textbf{C} & \textbf{R} & \textbf{U} & \textbf{T} & \cellcolor[HTML]{C0C0C0}{\textbf{M}} \\
        \midrule
        \midrule
        \rowcolor[HTML]{FFF9C0}
        \multicolumn{25}{l}{\textbf{\textit{CVQA}}} \\
        HCRN~\cite{le2020hierarchical} & - & - & - & - & 36.8 & - & - & - & - & - & - & - & - & - & - & - & 57.9 & - & - & - & - & - & 35.4 & - \\
        ClipBERT~\cite{lei2021less} & - & - & - & - & - & - & - & - & - & - & - & - & - & - & - & - & 60.3 & - &  - & - & - & - & 37.4 & - \\
        SiaSamRea~\cite{yu2021learning} & - & - & - & - & 45.5 & - & - & - & - & - & 39.8 & - & - & - & - & - & 60.2 & - & - & - & - & - & 41.6 & - \\
        MERLOT~\cite{zellers2021merlot} & - & - & - & - & - & - & - & - & - & - & 41.4 & - & - & - & - & - & \textbf{69.5} & - &  - & - & - & - & - & - \\
        All-in-one~\cite{wang2022all} & 62.6 & 31.5 & 4.5 & 0.0 & 42.8 & 7.9 & 65.1 & 34.1 & 6.9 & 0.0 & 39.5 & 5.3 & 79.4 & 34.5 & 5.7 & 0.0 &  65.6 & 10.1 & 50.4 & 12.3 & 0.8 & 0.0 & 39.5 & 3.9 \\
        JustAsk~\cite{yang2021just} & 65.9 & 37.8 & 13.6 & 0.0 & 47.5 & 12.6 & 60.5 & 37.1 & 16.9 & 0.0 & 39.0 & 8.2 & 68.0 & 31.3 & 11.4 & 0.0 & 56.9 & 11.7 & 51.7 & 18.5 & 6.0 & 0.0 & 41.8 & 7.0 \\
        VIOLET~\cite{fu2021violet} & \textbf{77.5} & 10.5 & 0.0 & 0.0 & 43.6 & 2.7 & 63.5 & 32.2 & 0.5 & 0.0 & 37.6 & 3.7 & \textbf{89.0} & 14.3 & 0.0 & 0.0 & 68.0 & 4.5 & 55.0 & 0.6 & 0.0 & 0.0 & 40.9 & 1.4 \\
        FrozenBiLM~\cite{yang2022zero} & 72.7 & \textbf{48.3} & 18.9 & 0.0 & 54.9 & 17.2 & 68.1 & \textbf{40.8} & 16.4 & 0.0 & 43.5 & 7.9 & 77.9 & 51.8 & 24.7 & 0.0 & 68.6 & 23.5 & \textbf{57.0} & 25.5 & 0.0 & 0.0 & 46.6 & 6.7 \\
        \midrule
        \rowcolor[HTML]{FFF9C0}
        \multicolumn{25}{l}{\textbf{\textit{OVQA}}} \\
        \textbf{All-in-one+} & \cellcolor[HTML]{BFF2FF}{62.8} & \cellcolor[HTML]{BFF2FF}{34.0} & \cellcolor[HTML]{BFF2FF}{6.3} & \cellcolor[HTML]{BFF2FF}{0.4} & \cellcolor[HTML]{BFF2FF}{43.8} & \cellcolor[HTML]{BFF2FF}{9.4} & \cellcolor[HTML]{FFD7D1}{64.9} & \cellcolor[HTML]{BFF2FF}{35.9} & \cellcolor[HTML]{BFF2FF}{9.8} & \cellcolor[HTML]{BFF2FF}{0.5} & \cellcolor[HTML]{BFF2FF}{40.2} & \cellcolor[HTML]{BFF2FF}{6.8} & \cellcolor[HTML]{FFD7D1}{78.3} & \cellcolor[HTML]{BFF2FF}{39.3} & \cellcolor[HTML]{BFF2FF}{10.2} & \cellcolor[HTML]{BFF2FF}{0.4} & \cellcolor[HTML]{BFF2FF}{66.0} & \cellcolor[HTML]{BFF2FF}{13.2} & \cellcolor[HTML]{FFD7D1}{49.8} & \cellcolor[HTML]{BFF2FF}{14.6} & \cellcolor[HTML]{BFF2FF}{1.6} & 0.0 & 39.5 & \cellcolor[HTML]{BFF2FF}{4.7} \\
        \textbf{JustAsk+} & \cellcolor[HTML]{FFD7D1}{65.6} & \cellcolor[HTML]{BFF2FF}{37.9} & 13.6 & \cellcolor[HTML]{BFF2FF}{6.3} & \cellcolor[HTML]{BFF2FF}{47.7} & \cellcolor[HTML]{BFF2FF}{14.5} & \cellcolor[HTML]{BFF2FF}{60.6} & 37.1 & \cellcolor[HTML]{FFD7D1}{16.7} & \cellcolor[HTML]{BFF2FF}{4.8} & \cellcolor[HTML]{BFF2FF}{40.0} & \cellcolor[HTML]{BFF2FF}{11.5} & 68.0 & \cellcolor[HTML]{BFF2FF}{32.1} & \cellcolor[HTML]{BFF2FF}{12.4} & \cellcolor[HTML]{BFF2FF}{9.8} & \cellcolor[HTML]{BFF2FF}{57.4} & \cellcolor[HTML]{BFF2FF}{14.4} & \cellcolor[HTML]{FFD7D1}{51.5} & \cellcolor[HTML]{FFD7D1}{18.4} & 6.0 & \cellcolor[HTML]{BFF2FF}{2.6} & 41.8 & \cellcolor[HTML]{BFF2FF}{7.6} \\
        \textbf{VIOLET+} & \cellcolor[HTML]{FFD7D1}{70.6} & \cellcolor[HTML]{BFF2FF}{38.8} & \cellcolor[HTML]{BFF2FF}{6.7} & \cellcolor[HTML]{BFF2FF}{0.1} & \cellcolor[HTML]{BFF2FF}{49.5} & \cellcolor[HTML]{BFF2FF}{10.7} & \cellcolor[HTML]{FFD7D1}{63.4} & \cellcolor[HTML]{BFF2FF}{37.1} & \cellcolor[HTML]{BFF2FF}{9.2} & \cellcolor[HTML]{BFF2FF}{0.6} & \cellcolor[HTML]{BFF2FF}{39.7} & \cellcolor[HTML]{BFF2FF}{6.1} & \cellcolor[HTML]{FFD7D1}{77.3} & \cellcolor[HTML]{BFF2FF}{38.9} & \cellcolor[HTML]{BFF2FF}{10.8} & \cellcolor[HTML]{BFF2FF}{2.0} &  \cellcolor[HTML]{FFD7D1}{65.3} & \cellcolor[HTML]{BFF2FF}{14.3} & \cellcolor[HTML]{FFD7D1}{53.8} & \cellcolor[HTML]{BFF2FF}{14.7} & \cellcolor[HTML]{BFF2FF}{0.9} & 0.0 & \cellcolor[HTML]{BFF2FF}{42.4} & \cellcolor[HTML]{BFF2FF}{4.5} \\
        \textbf{FrozenBiLM+}  & \cellcolor[HTML]{FFD7D1}{72.2} & \cellcolor[HTML]{FFD7D1}{48.2} & \textbf{\cellcolor[HTML]{BFF2FF}{21.6}} & \textbf{\cellcolor[HTML]{BFF2FF}{16.1}} & \textbf{\cellcolor[HTML]{BFF2FF}{55.8}} & \textbf{\cellcolor[HTML]{BFF2FF}{21.7}} & \textbf{\cellcolor[HTML]{BFF2FF}{68.8}} & \cellcolor[HTML]{FFD7D1}{39.9} & \textbf{\cellcolor[HTML]{BFF2FF}{17.3}} & \textbf{\cellcolor[HTML]{BFF2FF}{5.8}} & \textbf{\cellcolor[HTML]{BFF2FF}{44.8}} & \textbf{\cellcolor[HTML]{BFF2FF}{12.4}} & \cellcolor[HTML]{FFD7D1}{77.7} & \textbf{\cellcolor[HTML]{BFF2FF}{52.1}} & \textbf{\cellcolor[HTML]{BFF2FF}{28.6}} & \textbf{\cellcolor[HTML]{BFF2FF}{21.3}} & \cellcolor[HTML]{BFF2FF}{69.0} & \textbf{\cellcolor[HTML]{BFF2FF}{30.2}} & \cellcolor[HTML]{FFD7D1}{56.1} & \textbf{\cellcolor[HTML]{BFF2FF}{26.6}} & \textbf{\cellcolor[HTML]{BFF2FF}{11.7}} & \textbf{\cellcolor[HTML]{BFF2FF}{6.6}} & \textbf{\cellcolor[HTML]{BFF2FF}{47.0}} & \textbf{\cellcolor[HTML]{BFF2FF}{12.4}} \\ 
        \bottomrule
    \end{tabular}
    \end{adjustbox}
    \caption{\textbf{Comparison with state-of-the-art models.}
    B, C, R, U, T, and M refer to Base, Common, Rare, Unseen, Total, and mean accuracy (mAcc), respectively.
    + denotes our developed version of baselines for OVQA.
    \textcolor{blue}{Blue} cell denotes performance increase and \textcolor{red}{red} cell denotes performance decrease compared to the baselines.
    }
    \label{tab:main}
\end{table*}

\subsection{Label smoothing}

After constructing the answer graph, we extract answer embeddings $V_\text{train} = \{v_i\}_{i=1}^{|\mathcal{V}_\text{train}^{(K)}|}\in \mathbb{R}^{|\mathcal{V}_\text{train}^{(K)}| \times D}$ and $V_\text{test} = \{v_i\}_{i=1}^{|\mathcal{V}_\text{test}^{(K)}|} \in \mathbb{R}^{|\mathcal{V}_\text{test}^{(K)}| \times D}$ using the answer encoder (\textit{e.g.}, DeBERTa tokenizer) and they are used as input node features, \textit{i.e.}, $\mathbf{h}_i^{(0)}$ in Eq.~\eqref{eq:message} is $v_i$.
Note that the answer encoder is frozen during training.
At the training phase, a node feature $V_\text{train}$ and a graph structure $\mathcal{G}_\text{train}^{(K)}$ are fed into a GNN.

As for a message-passing algorithm, we modify the standard graph attention network (GAT) to adopt the attention mechanism and use it to adjust the information taken from the neighbor nodes.
The attention score from the $j$-th to $i$-th node is calculated as:
\begin{equation}
    \resizebox{1.0\hsize}{!}{
    $\alpha_{ij}^{(l)} = 
    \frac{\exp \left(\text{LeakyReLU}
    \left(\left({\mathbf{W}^{(l)}_\text{dst}}\mathbf{h}_i^{(l-1)}\right)^\top
    \left(\mathbf{W}^{(l)}_\text{src}\mathbf{h}_j^{(l-1)}\right)\right)\right)}
    {\sum\limits_{k \in \mathcal{N}_i}\exp \left(\text{LeakyReLU}
    \left(\left({\mathbf{W}^{(l)}_\text{dst}}\mathbf{h}_i^{(l-1)}\right)^\top
    \left(\mathbf{W}^{(l)}_\text{src}\mathbf{h}_k^{(l-1)}\right)\right)\right)}
    $},
    \label{eq:attention}
\end{equation}
where $\mathbf{W}^{(l)}_\text{src} \in \mathbb{R}^{D \times D}$ and $\mathbf{W}^{(l)}_\text{dst} \in \mathbb{R}^{D \times D}$ are learnable weight matrices to project source and destination node features, respectively.
In Eq.~\eqref{eq:attention}, the attention score $\alpha_{ij}^{(l)}$ is computed based on the similarity between source node $j$ and target node $i$.
Subsequently, AGGREGATE function in Eq.~\eqref{eq:message} is defined as:
\begin{equation}
    \text{AGGREGATE}\left(\mathbf{h}_j^{(l-1)}: j \in \mathcal{N}_i\right) \triangleq \sum_{j \in \mathcal{N}_i} \alpha_{ij}^{(l)} \mathbf{h}_j^{(l-1)},
    \label{eq:aggregate}
\end{equation}
the weighted sum of neighbor node features based on the attention score $\alpha_{ij}^{(l)}$.

After $L$-layer GNN, the output answer embeddings are obtained as $\mathbf{H}_\text{train} = [\mathbf{h}_1^{(L)}, \mathbf{h}_2^{(L)}, \dots, \mathbf{h}_i^{(L)}, \dots]^\top \in \mathbb{R}^{|\mathcal{V}_\text{train}| \times D}$, where $\forall_i \in \mathcal{V}_\text{train}$.
We use two layer GNNs, \textit{i.e.}, $L = 2$, to aggregate the information up to 2-hop neighborhoods.
For learning stability, we adopt convex combinations of output answer embeddings of a GNN-based soft verbalizer, $\mathbf{H}_\text{train}$, with input answer embeddings $V_\text{train}$ as:
\begin{equation}
    \hat{\mathbf{H}}_\text{train} = \varepsilon \cdot V_\text{train} + (1 - \varepsilon) \cdot \mathbf{H}_\text{train},
    \label{eq:epsilon}
\end{equation}
where $\varepsilon$ is a convex combination coefficient.
Also, we fix the weight matrix $\mathbf{W}^{(l)}$ in Eq.~\eqref{eq:message} of the main paper to an identity matrix.
Stop-gradient is applied to the input answer embeddings (\textit{i.e.}, frozen answer encoder) so the additional trainable parameters in GNN-based soft verbalizer are $\mathbf{W}^{(l)}_\text{src}$ and $\mathbf{W}^{(l)}_\text{dst}$ in Eq.~\eqref{eq:attention}.

Finally, the similarity is calculated between the output feature of [MASK] token of the cross-modal encoder, $\mathbf{m}$, and the smoothed answer embeddings $\hat{\mathbf{H}}_\text{train}$ to predict the label, \textit{i.e.}, $\hat{\mathbf{H}}_\text{train} \mathbf{m} \in \mathbb{R}^{|\mathcal{V}_\text{train}|}$.
Both GNN and backbone architectures are trained with the following loss:
\begin{equation}
    \mathcal{L} = \text{CrossEntropy}\left(a_\text{GT}, \text{Softmax}\left(\hat{\mathbf{H}}_\text{train} \mathbf{m}\right) \right),
    \label{eq:loss}
\end{equation}
where $a_\text{GT}$ is a ground-truth answer.
During training, our GNN-based soft verbalizer learns to smooth the original answers with their neighborhoods.
In the test phase, the learned smoothing function softly updates information from their neighborhoods for the test vocabulary that includes rare and unseen answers.
As a result, the GNN-based soft verbalizer enhances prediction on the out-of-vocabulary answers and alleviates the strong bias toward the frequent answers.
\section{Experiments}

\begin{table*}[!ht]
    \centering
    \setlength{\tabcolsep}{3.5pt}
    \begin{adjustbox}{width=\linewidth}
    \begin{tabular}{c|c|c c c c c c|c c c c c c|c c c c c c|c c c c c c}
        \toprule
        \textbf{Models} & \textbf{GNN-based} & \multicolumn{6}{c|}{\textbf{MSVD-QA}} & \multicolumn{6}{c|}{\textbf{ActivityNet-QA}} & \multicolumn{6}{c|}{\textbf{TGIF-QA}} & \multicolumn{6}{c}{\textbf{MSRVTT-QA}} \\
        & \textbf{soft verbalizer} & \textbf{B} & \textbf{C} & \textbf{R} & \textbf{U} & \textbf{T} & \cellcolor[HTML]{C0C0C0}{\textbf{M}} & \textbf{B} & \textbf{C} & \textbf{R} & \textbf{U} & \textbf{T} & \cellcolor[HTML]{C0C0C0}{\textbf{M}} & \textbf{B} & \textbf{C} & \textbf{R} & \textbf{U} & \textbf{T} & \cellcolor[HTML]{C0C0C0}{\textbf{M}} & \textbf{B} & \textbf{C} & \textbf{R} & \textbf{U} & \textbf{T} & \cellcolor[HTML]{C0C0C0}{\textbf{M}} \\
        \midrule
        \midrule
        \multirow{2}{*}{FrozenBiLM+} & \ding{56} & 72.1 & 47.8 & 20.3 & 13.7 & 55.4 & 20.8 & 67.7 & 37.4 & 15.5 & 4.2 & 43.2 & 10.4 & 77.5 & 51.7 & 28.5 & 18.7 & 68.9 & 30.1 & 55.8 & 26.4 & 11.4 & 5.8 & 46.7 & 12.1 \\ 
        & \ding{52} & \textbf{72.2} & \textbf{48.2} & \textbf{21.6} & \textbf{16.1} & \textbf{55.8} & \textbf{21.7} & \textbf{68.8} & \textbf{39.9} & \textbf{17.3} & \textbf{5.8} & \textbf{44.8} & \textbf{12.4} & \textbf{77.7} & \textbf{52.1} & \textbf{28.6} & \textbf{21.3} & \textbf{69.0} & \textbf{30.2} & \textbf{56.1} & \textbf{26.6} & \textbf{11.7} & \textbf{6.6} & \textbf{47.0} & \textbf{12.4} \\
        \bottomrule
    \end{tabular}
    \end{adjustbox}
    \caption{\textbf{Effectiveness of GNN-based soft verbalizer on various datasets}
    }
    \label{tab:ablation}
\end{table*}
\subsection{Experimental setup}

\noindent \textbf{Datasets and answer vocabularies.}
Our experiment covers four open-ended VideoQA datasets: MSVD-QA~\cite{xu2017video}, MSRVTT-QA~\cite{xu2017video}, ActivityNet-QA~\cite{yu2019activitynet}, and TGIF-FrameQA~\cite{jang2017tgif}. 
For training/testing, MSVD-QA is split into 32K/13K. 
MSRVTT-QA follows 159K/73K. 
ActivityNet-QA splits into 32K/8K. 
TGIF-FrameQA uses 39K/13K.
The specific numbers of train/test vocabularies respectively for each dataset are as follows: MSVD-QA 1852/1200, MSRVTT-QA 4000/4173, TGIF-FrameQA 1540/933, and ActivityNet-QA 1654/2103.

\noindent \textbf{Baselines}
We introduce new baselines by modifying existing open-ended VideoQA models: All-in-one~\cite{wang2022all}, JustAsk~\cite{yang2021just}, VIOLET~\cite{fu2021violet}, and FrozenBiLM~\cite{yang2022zero}.
We follow the vocabulary setting of each baseline to reproduce their performances.

\noindent \textbf{Implementation details.}
We adopt GloVe~\cite{pennington2014glove} as an extra knowledge base to construct the answer graph.
We use nearest neighborhood words of the original answer based on GloVe word embeddings to create the neighbor nodes.
The answer graph is constructed by considering up to 2-hop neighborhoods from the original answer.
We search $\varepsilon$ in $\{0.5, 0.6, 0.7, 0.8, 0.9\}$.
Further dataset and implementation details for baselines are provided in the supplement.
\begin{table}[t!]
    \centering
    \setlength{\tabcolsep}{3.5pt}
    \begin{adjustbox}{width=\linewidth}
    \begin{tabular}{c|c|c c c c c >{\columncolor{lightgray}}c}
        \toprule
        \textbf{Models} & \textbf{GNN-based} & \multicolumn{6}{c}{\textbf{ActivityNet}} \\
         & \textbf{soft verbalizer} & \textbf{B} & \textbf{C} & \textbf{R} & \textbf{U} & \textbf{T} & \textbf{M} \\
        \midrule
        \multirow{2}{*}{All-in-one+} & \ding{56} & 64.9 & 35.9 & 9.8 & 0.5 & 40.2 & 6.8 \\ 
        & \ding{52} & \textbf{65.0} & \textbf{40.8} & \textbf{13.8} & \textbf{1.6} & \textbf{42.0} & \textbf{8.7} \\
        \midrule
        \multirow{2}{*}{JustAsk+} & \ding{56} & 60.6 & \textbf{37.1} & 16.7 & 4.8 & 40.0 & 11.5 \\ 
        & \ding{52} & \textbf{61.5} & 35.6 & \textbf{18.9} & \textbf{5.1} & \textbf{40.4} & \textbf{12.1} \\
        \midrule
        \multirow{2}{*}{VIOLET+} & \ding{56} & 63.4 & \textbf{37.1} & 9.2 & \textbf{0.6} & 39.7 & 6.1 \\
        & \ding{52} & \textbf{63.6} & 36.1 & \textbf{12.9} & \textbf{0.6} & \textbf{39.9} & \textbf{7.4} \\
        \bottomrule
    \end{tabular}
    \end{adjustbox}
    \caption{\textbf{Effectiveness of GNN-based soft verbalizer on various backbone models.}
    }
    \label{tab:baselines}
\end{table}
\subsection{Evaluation on OVQA}

We first evaluate the open-ended VideoQA baseline models under both settings of CVQA and OVQA.
In OVQA, we additionally introduce an answer encoder, DeBERTa~\cite{he2020deberta} tokenizer, to extract the answer embeddings.
In Tab.~\ref{tab:main}, for all the previous models in CVQA in general, the total performance (\textbf{T}) seems plausible but mAcc (\textbf{M}) is extremely low, \textit{e.g.}, the total performance (\textbf{T}) of VIOLET is 40.9\% but the accuracy of the non-base answers (\textbf{C}, \textbf{R}, \textbf{U}) is almost 0\% resulting in 1.4\% mAcc (\textbf{M}) on MSRVTT-QA.
This means that previous CVQA baselines are highly biased toward frequent answers and fail to generalize on rare and unseen answers.

On the other hand, by comparing Baseline (CVQA) and Baseline+ (OVQA) over the four baselines, mAcc (\textbf{M}) of OVQA baselines are impressively increased on all datasets.
In detail, mAcc (\textbf{M}) of FrozenBiLM+ is improved by 4.5\%, 4.5\%, 6.7\%, and 5.7\% compared to FrozenBiLM on each dataset.
As for the detailed accuracy of each category, the performance on base answers (\textbf{B}) tends to marginally decrease, but the performance on others including the total performance significantly increases.
This result indicates that further taking into account non-frequent answers is beneficial for total performance as well as mAcc.
We also observe that baselines equipped with language models (\textit{e.g.}, JustAsk with DistillBERT~\cite{sanh2019distilbert} and FrozenBiLM with DeBERTa~\cite{he2020deberta}) show relatively larger improvement in unseen answers (\textbf{U}).

The gap between the total performances (\textbf{T}) of standard VIOLET and All-in-one is 0.8\% on MSVD-QA.
Specifically, the performance of base (\textbf{B}) and common answers (\textbf{C}) are 77.5\% and 10.5\% on VIOLET and 62.6\% and 31.5\% on All-in-one, respectively.
This demonstrates that VIOLET is more biased toward base answers than All-in-one while the total performance is similar.
This is also shown by comparing their mAcc (\textbf{M}) (7.9\% on All-in-one but 2.7\% on VIOLET).
Interestingly, our variant VIOLET+ significantly outperforms the standard VIOLET by a large margin of 5.9\% and 8\% in terms of the total performance (\textbf{T}) and mAcc (\textbf{M}) on MSVD-QA, respectively.
The performance gain mainly comes from the common answers (\textbf{C}) while being improved from 10.5\% to 38.8\%.
On the other hand, the total performance gap between All-in-one and All-in-one+ is relatively smaller than VIOLET, implying that the performance gain is significant if the model is highly biased toward base (frequent) answers.
\subsection{Ablation studies on GNN-based soft verbalizer}
\label{subsec:ablation}

\noindent \textbf{Effectiveness of GNN-based soft verbalizer.}
In Tab.~\ref{tab:ablation}, we conduct the ablation study of GNN-based soft verbalizer on FrozenBiLM+.
By comparing FrozenBiLM+ with and without GNN-based soft verbalizer, the performance gains of unseen answers (\textbf{U}) are 2.4\%, 1.6\%, 2.6\%, and 0.8\% on MSVD-QA, ActivityNet-QA, TGIF-QA, and MSRVTT-QA respectively.
The performances on base and common answers (\textbf{B}, \textbf{C}) are also improved across all datasets implying that GNN-based soft verbalizer is beneficial to not only rare and unseen answers but also base and common answers.

\begin{figure}[t] 
    \vspace{-2mm}
    \centering
    \begin{subfigure}[t]{0.45\linewidth}
        \includegraphics[width=1.0\linewidth]{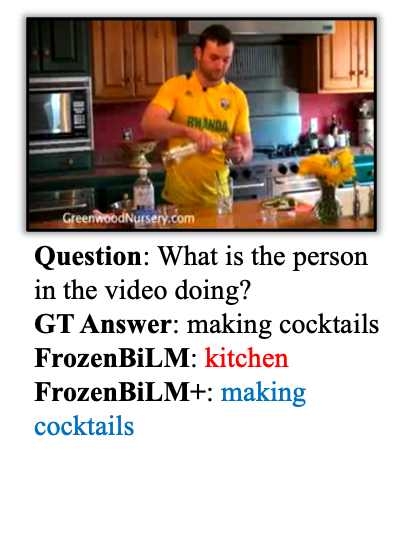}        
        \vspace{-9mm}
        \caption{ }
        \label{fig:qualitative1}
    \end{subfigure}
    \begin{subfigure}[t]{0.45\linewidth}
        \includegraphics[width=1.0\linewidth]{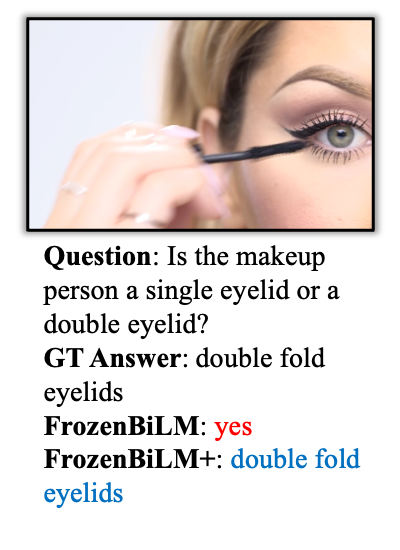}
        \vspace{-9mm}
        \caption{ }
        \vspace{3mm}
        \label{fig:qualitative2}
    \end{subfigure}
    \begin{subfigure}[t]{0.45\linewidth}
        \includegraphics[width=1.0\linewidth]{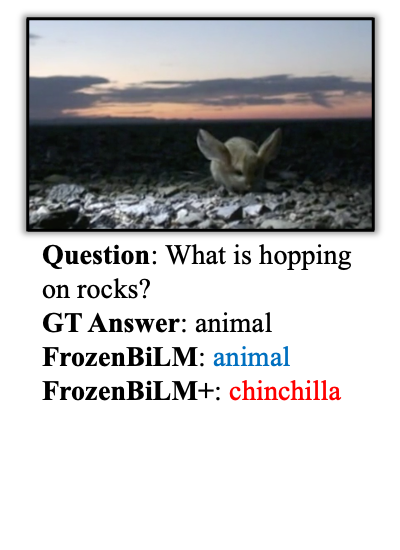}
        \vspace{-18mm}
        \caption{ }
        \label{fig:qualitative3}
    \end{subfigure}
    \begin{subfigure}[t]{0.45\linewidth}
        \includegraphics[width=1.0\linewidth]{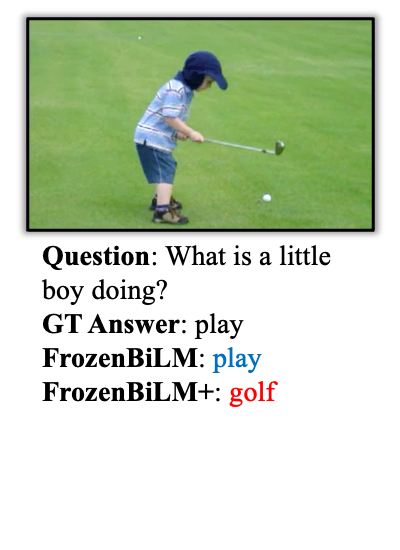}
        \vspace{-18mm}
        \caption{ }
        \label{fig:qualitative4}
    \end{subfigure}
    \vspace{-8mm}
    \caption{\textbf{Examples of unseen answers.}
    (a) and (b) are success cases and (c) and (d) are failure cases.
    }
    \label{fig:qualitative}
\end{figure}

Furthermore, the performance gain of base and common answers (\textbf{B}, \textbf{C}) is larger on AcitivityNet-QA than other datasets.
We conjecture that this comes from the dataset annotations where most unseen answers on datasets except for ActivityNet-QA consist of hyponyms of base and common answers.
For example, in MSVD-QA, `play’ (hypernym) is in base answers while `golf’ (hyponym) belongs to unseen answers.
GNN-based soft verbalizer enables the model to accurately predict the answer `golf' yet according to the annotation, the ground-truth answer is `play' (See Fig.~\ref{fig:qualitative4} for details).
Hence, this sometimes leads to the performance degradation on base answers by trying to predict accurate hyponym.
On the other hand, most unseen answers in ActivityNet-QA comprise phrases that cannot be covered by base answers like `double fold eyelids' (Fig.~\ref{fig:qualitative2}), and thus considering unseen answers does not affect the performance on base answers.
As a result, the performances on base and common answers are also increased by a large margin along with the improvements on rare and unseen answers.

Tab.~\ref{tab:baselines} also shows the effectiveness of GNN-based soft verbalizer by applying it to various backbone models.
We extract answer embeddings in an offline manner using frozen answer encoder (DeBERTa tokenizer) on All-in-one and VIOLET.
On the other hand, JustAsk uses its own answer encoder which is unfrozen during training so we adopt a 2-stage training scheme: train the answer encoder of JustAsk first and then train our GNN-based soft verbalizer with the trained answer encoder frozen.
With a GNN-based soft verbalizer, the total performance (\textbf{T}) and mAcc (\textbf{M}) are consistently improved on all other models.
Especially, the performances of rare answers (\textbf{R}) are increased by 4\%, 2.2\%, and 3.7\% on All-in-one+, JustAsk+, and VIOLET+, signifying that GNN-based soft verbalizer is a generally applicable algorithm.

\begin{table}[t!]
    \centering
    \setlength{\tabcolsep}{3.5pt}
    \begin{adjustbox}{width=\linewidth}
    \begin{tabular}{c|c|c|c c c c c >{\columncolor{lightgray}}c}
        \toprule
        & \multicolumn{2}{c|}{\textbf{Verbalizer}} & \multicolumn{6}{c}{\textbf{ActivityNet}} \\
        & \textbf{Answer graph} & \textbf{soft/hard} & \textbf{B} & \textbf{C} & \textbf{R} & \textbf{U} & \textbf{T} & \textbf{M} \\
        \midrule
        \midrule
        (A) & \multicolumn{2}{c|}{N/A} & 67.7 & 37.4 & 15.5 & 4.2 & 43.2 & 10.4 \\
        \midrule
        (B) & \ding{56} & hard & 68.1 & 31.0 & 10.2 & 3.0 & 41.2 & 7.9\\
        (C) & \ding{56} & soft & \textbf{68.9} & 39.1 & 16.7 & 4.7 & 44.4 & 10.8 \\
        (D) & \ding{52} & hard & 68.3 & 37.6 & 15.4 & 4.5 & 43.6 & 10.5\\ 
        (E) & \ding{52} & soft & 68.8 & \textbf{39.9} & \textbf{17.3} & \textbf{5.8} &    \textbf{44.8} & \textbf{12.4} \\
        \bottomrule
    \end{tabular}
    \end{adjustbox}
    \caption{\textbf{Comparison of each verbalizer type on FrozenBiLM+.}
    (A) does not adopt the verbalizer. 
    (B) uses neither answer graph nor learnable verbalizer, \textit{i.e.}, only conducting mean-pooling of similar words from the external knowledge base.
    (C) adapts an MLP to be trainable from (B).
    Both (D) and (E) construct answer graph but (D) uses the mean-pooled feature of fixed answer embeddings while (E) adaptively adjusts them.
    Note that (E) is our GNN-based soft verbalizer.
    }
    \label{tab:verbalizer}
\end{table}
\noindent \textbf{Comparison of various verbalizers.}
We also compare various verbalizers with our GNN-based soft verbalizer in Tab.~\ref{tab:verbalizer}.
First, the method with a hard verbalizer (B), which utilizes a mean-pooled feature of similar words from the external knowledge base, exhibits considerable
degradation compared to the method without a verbalizer (A).
However, (C) outperforms both (A) and (B) demonstrating that leveraging a soft verbalizer with a learnable MLP layer improves the model performance by adequately adjusting the information of similar words.
Also in general, (D) and (E) surpass (B) and (C), respectively, indicating that constructing the verbalizer with answer graphs and message-passing algorithms leads to more effective answer embeddings.
Specifically, our full model (E) outperforms (C) by 0.6\% and 1.1\% for rare and unseen respectively resulting in 1.6\% improvement in mAcc.
This demonstrates that our GNN-based soft verbalizer adaptively aggregates the information of similar words on answer graphs and yields more effective answer embeddings.
\subsection{Qualitative results}

\noindent \textbf{Examples of unseen answers.}
Fig.~\ref{fig:qualitative} shows qualitative results on the unseen answers comparing FrozenBiLM and our FrozenBiLM+.
For example in Fig.~\ref{fig:qualitative1}, FrozenBiLM is limited to the answer only within the closed-vocabulary set, ``kitchen'', for the question ``What is the person in the video doing?''. 
On the other hand, FrozenBiLM+ is capable of predicting the out-of-vocabulary answer ``making cocktails'' with the guidance of answer embeddings from the answer encoder. 
Furthermore, FrozenBiLM is biased toward frequent answers by considering only \textit{top-k} candidates.
Specifically on ActivityNet-QA (Fig.~\ref{fig:qualitative2}), it tends to predict ``yes'' on the question starting with ``Is'' since 97\% of answers to such question types are ``yes'' or ``no''.
This language bias is commonly observed in question answering tasks ~\cite{niu2021counterfactual, ramakrishnan2018overcoming, cadene2019rubi}. 
However, unlike the baseline, our model alleviates such bias and corrects the output to ``double fold eyelids''.
Finally, Fig.~\ref{fig:qualitative3} illustrates the failure case when the unseen answer is considered in MSVD-QA.
As mentioned in Sec.~\ref{subsec:ablation}, since most unseen answers are hyponyms of base and common answers, accurately predicting the answer as `chinchilla' is regarded as incorrect although the visual content actually depicts `chinchilla'.

\begin{figure}[t] 
    \centering
        \includegraphics[width=0.99\linewidth]{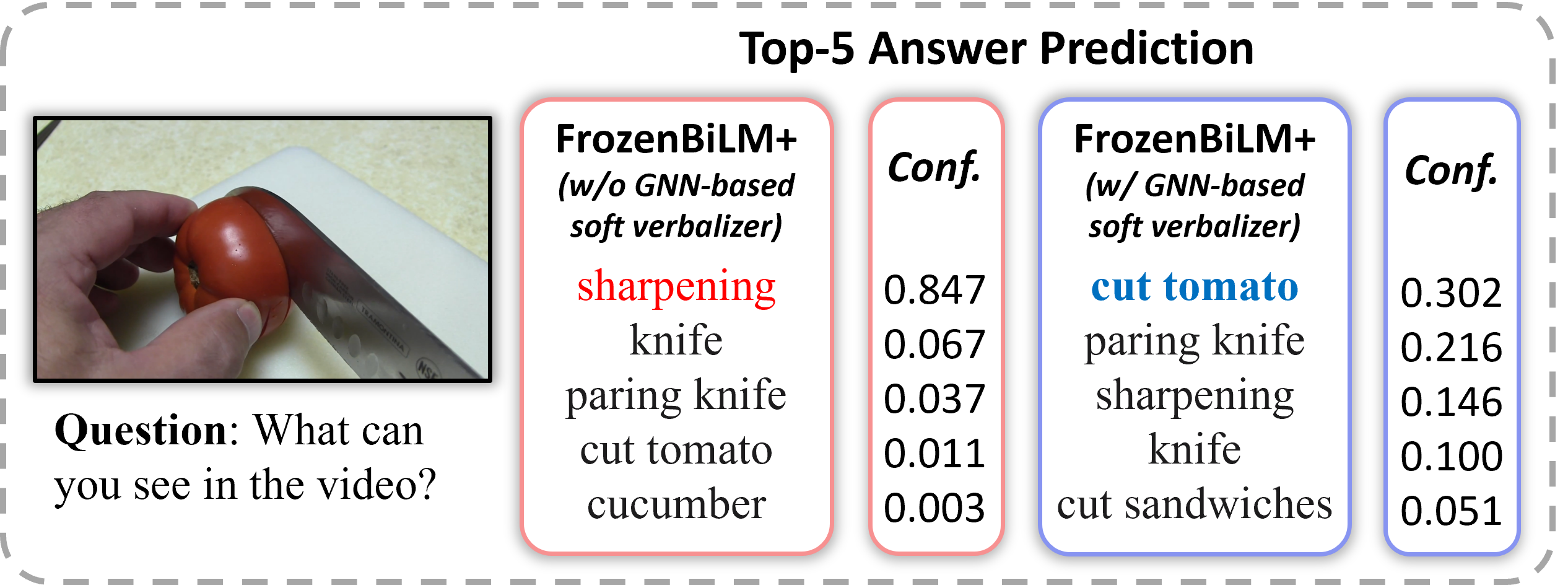}
        \caption{\textbf{Confidence scores of the top-5 predictions w/ and w/o GNN-based soft verbalizer on FrozenBiLM+.}}
        \label{fig:top5}
\end{figure}

\begin{figure}[t]
    \vspace{-1mm}
    \centering
    \includegraphics[width=\linewidth]{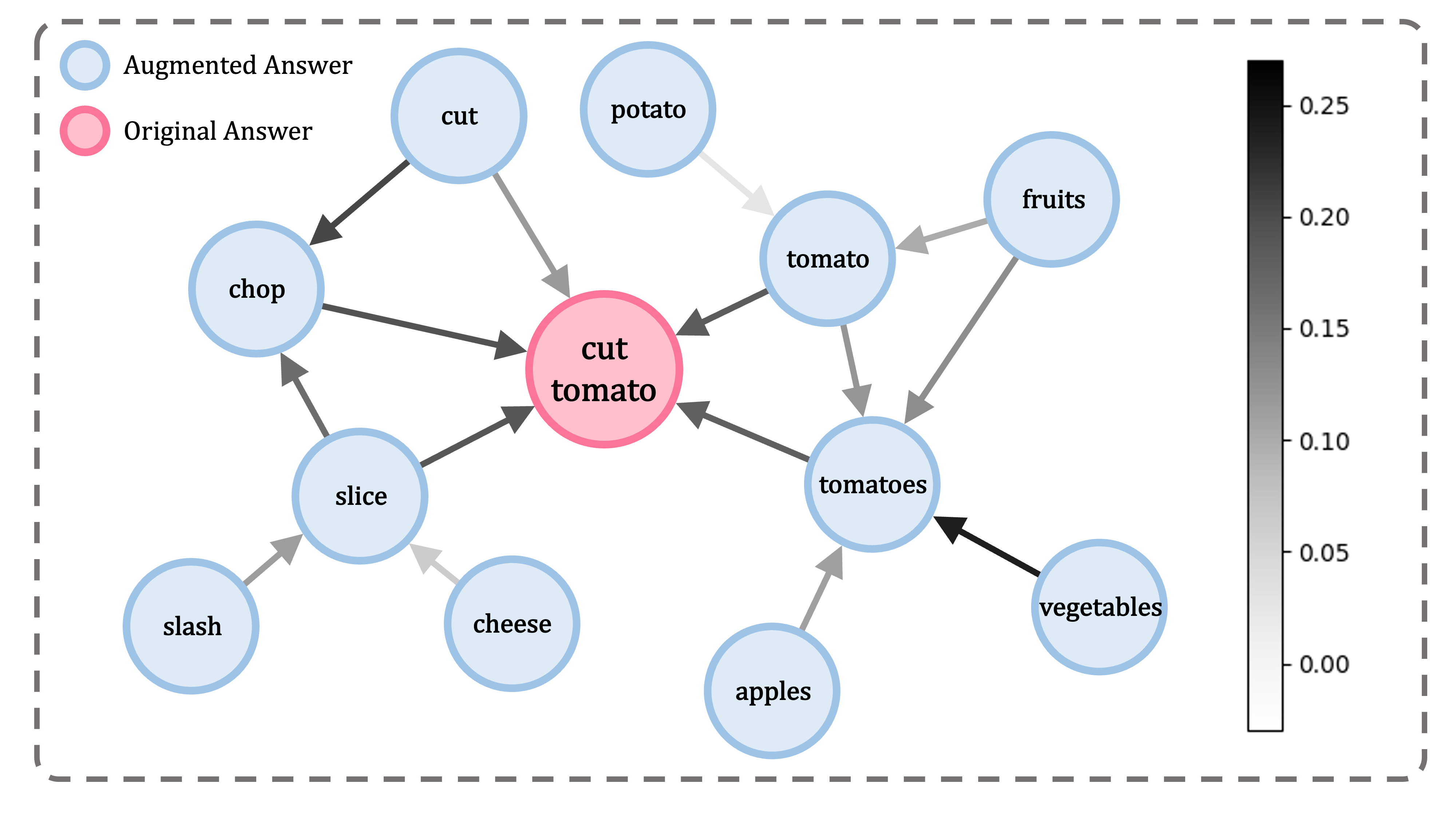}
    \caption{\textbf{Visualization of the attention score of our GNN, $\alpha_{ij}$, in terms of the answer ``cut tomato''.}
    The intensity of edges refers to the attention score $\alpha_{ij}$.
    }
    \vspace{-1mm}
    \label{fig:graph}
\end{figure}

\noindent \textbf{Visualization of GNN-based soft verbalizer.}
In Fig.~\ref{fig:top5}, we also qualitatively compare the models with and without a GNN-based soft verbalizer on FrozenBiLM+.
Without a GNN-based soft verbalizer, the model is over-confident in the wrong answer ``sharpening''.
However, with a GNN-based soft verbalizer, the model corrects its output to ``cut tomato'' regularizing its over-confidence.
To show how the GNN-based soft verbalizer smoothes the original answer, in Fig.~\ref{fig:graph}, we illustrate the attention score $\alpha_{ij}$ in Eq.~\eqref{eq:attention}.
We observe that GNN-based soft verbalizer aggregates the information mainly from ``chop'', ``slice'', and ``tomatoes'' to predict the answer ``cut tomato''.
On the other hand, it is reluctant to utilize the information of ``cheese'' or ``potato'', which are less relevant to the video, although they belong to the neighborhoods. 
This reveals that the answer embeddings are effectively updated by GNN-based soft verbalizer through adjusting the neighborhood information.

\section{Related works}

\noindent \textbf{Video question answering (VideoQA).}
VideoQA aims to align the dynamic visual contents with the linguistic semantics of a question to yield the answer.
The recent paradigm is to first pretrain the model on a vast amount of video-text paired data~\cite{zellers2021merlot, bain2021frozen, miech2019howto100m} and fine-tune it on VideoQA~\cite{wang2022all, fu2021violet,  yang2022zero, bain2021frozen, zeng2022x, li2022align}.
Typical VideoQA benchmarks take two formats: multiple-choice~\cite{li2020hero, lei2018tvqa} and open-ended~\cite{yang2021just, jang2017tgif, xu2017video, yu2019activitynet}.
In contrast to multiple-choice VideoQA where several answer options are provided for each question, the goal of open-ended VideoQA is to predict the answer without any candidate answers.
While existing open-ended VideoQA models~\cite{lei2021less,wang2022all,li2020hero,fu2021violet,zellers2021merlot,le2020hierarchical,yang2022zero} are promising, they still show sub-optimal performance due to the common practice of open-ended VideoQA that converts the task to a classification with only frequent answer candidates.
To alleviate such issues, we introduce a novel benchmark to incorporate open-vocabulary setting into the VideoQA model.

\noindent \textbf{Open-vocabulary visual understanding.}
The goal of open-vocabulary visual understanding is to predict arbitrary text categories not observed during model training.
There exist open-vocabulary classification models~\cite{radford2021learning, jia2021scaling} that leverage huge amounts of image-text pairs from the web and are trained with contrastive loss to make visual and language representations well aligned.
Recently, Open-Vocabulary Object Detection (OVOD)~\cite{zareian2021open,minderer2022simple,gu2021open, zhong2022regionclip, du2022learning, zhou2022detecting} has also gained attention, which targets to predict both base and unseen classes by training on a large-scale dataset that covers diverse vocabularies.
Also, open-vocabulary image segmentation ~\cite{huynh2022open,zhou2022maskclip,zhao2017open,xu2022simple,liang2022open,bucher2019zero,li2022language,ghiasi2022scaling,ding2022open} has arisen to localize unseen classes in a pixel level.
In this work, we extend this open-vocabulary setting to open-ended VideoQA to handle the out-of-vocabulary answers.
\section{Conclusion}

In this paper, we propose a new benchmark, Open-vocabulary Video Question Answering (OVQA), that evaluates the generalizability of the model for four different answer categories: base, common, rare, and unseen.
Moreover, we present a novel GNN-based soft verbalizer that smoothes label embeddings on answer graphs augmented with similar words from an external knowledge base to enhance prediction on out-of-vocabulary answers.
Evaluation of our developed baselines under the OVQA setting shows the merit of integrating an additional answer encoder that enables prediction on rare and unseen candidates. 
In addition, with extensive ablation studies and qualitative analyses, we validate the effectiveness of our GNN-based soft verbalizer in mitigating the bias of the model toward frequent answers and show the general applicability of the algorithm. 

\noindent \textbf{Acknowledgments.}
This work was partly supported by IITP grant funded by the Korea government (MSIT) (No.2022-0-01198), ICT Creative Consilience program (IITP-2023-2020-0-01819) supervised by the IITP, the National Supercomputing Center with supercomputing resources including technical support (KSC-2022-CRE-0261), and KakaoBrain corporation.
\newpage

{\small
\bibliographystyle{unsrt}
\bibliography{main.bbl}
}

\clearpage
\appendix
\section*{\Large Appendix}
\begin{figure}[t] 
    \centering
    \begin{subfigure}[t]{0.495\linewidth}
        \includegraphics[width=1.0\linewidth]{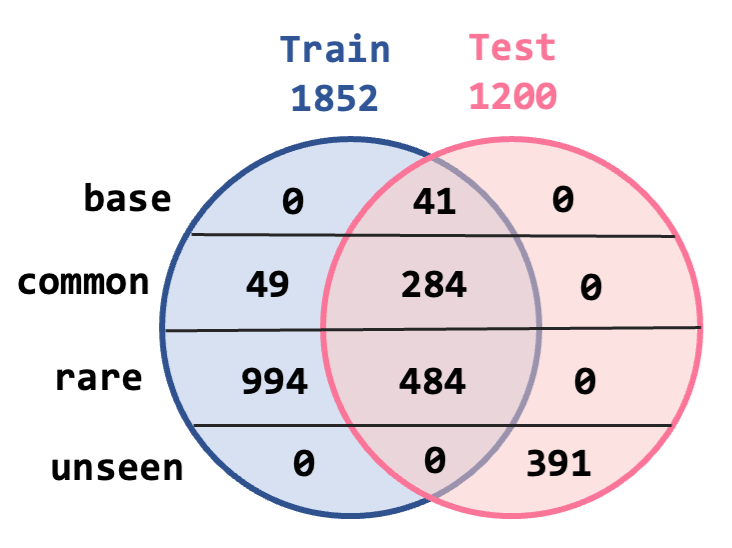}
        \caption{MSVD-QA}
        \label{fig:msvd_venn}
    \end{subfigure}
    \begin{subfigure}[t]{0.495\linewidth}
        \includegraphics[width=1.0\linewidth]{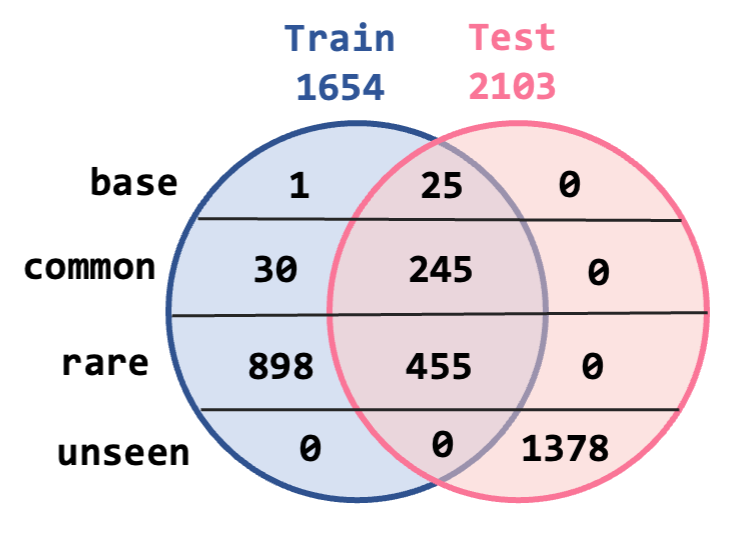}
        \caption{ActivityNet-QA}
        
        \label{fig:activitynet_venn}
    \end{subfigure}
        \begin{subfigure}[t]{0.495\linewidth}
        \includegraphics[width=1.0\linewidth]{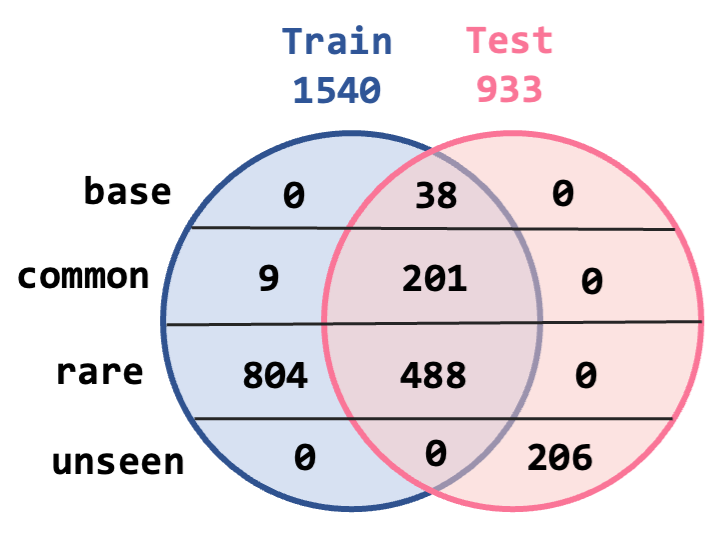}
        \caption{TGIF-QA}
        \label{fig:tgif_venn}
    \end{subfigure}
    \begin{subfigure}[t]{0.495\linewidth}
        \includegraphics[width=1.0\linewidth]{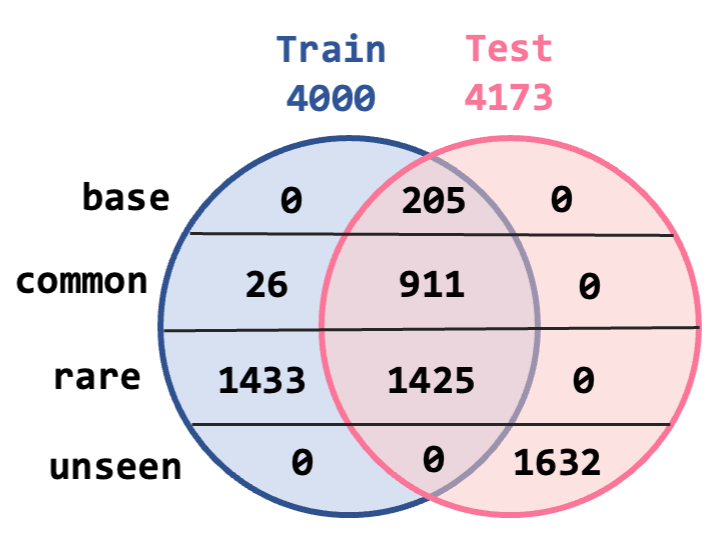}
        \caption{MSRVTT-QA}
        
        \label{fig:mstvtt_venn}
    \end{subfigure}
    \caption{\textbf{Dataset Venn diagram.}
     The distribution of rare, common, and frequent categories in train and test sets for four benchmark datasets. 
     The total number of vocabularies for each set is specified under the corresponding title.
    }
    \label{fig:venn_diagram}
\end{figure}
\section{Dataset details}
\begin{figure}[t] 
    \centering
    \begin{subfigure}[t]{0.45\linewidth}
        \includegraphics[width=1.0\linewidth]{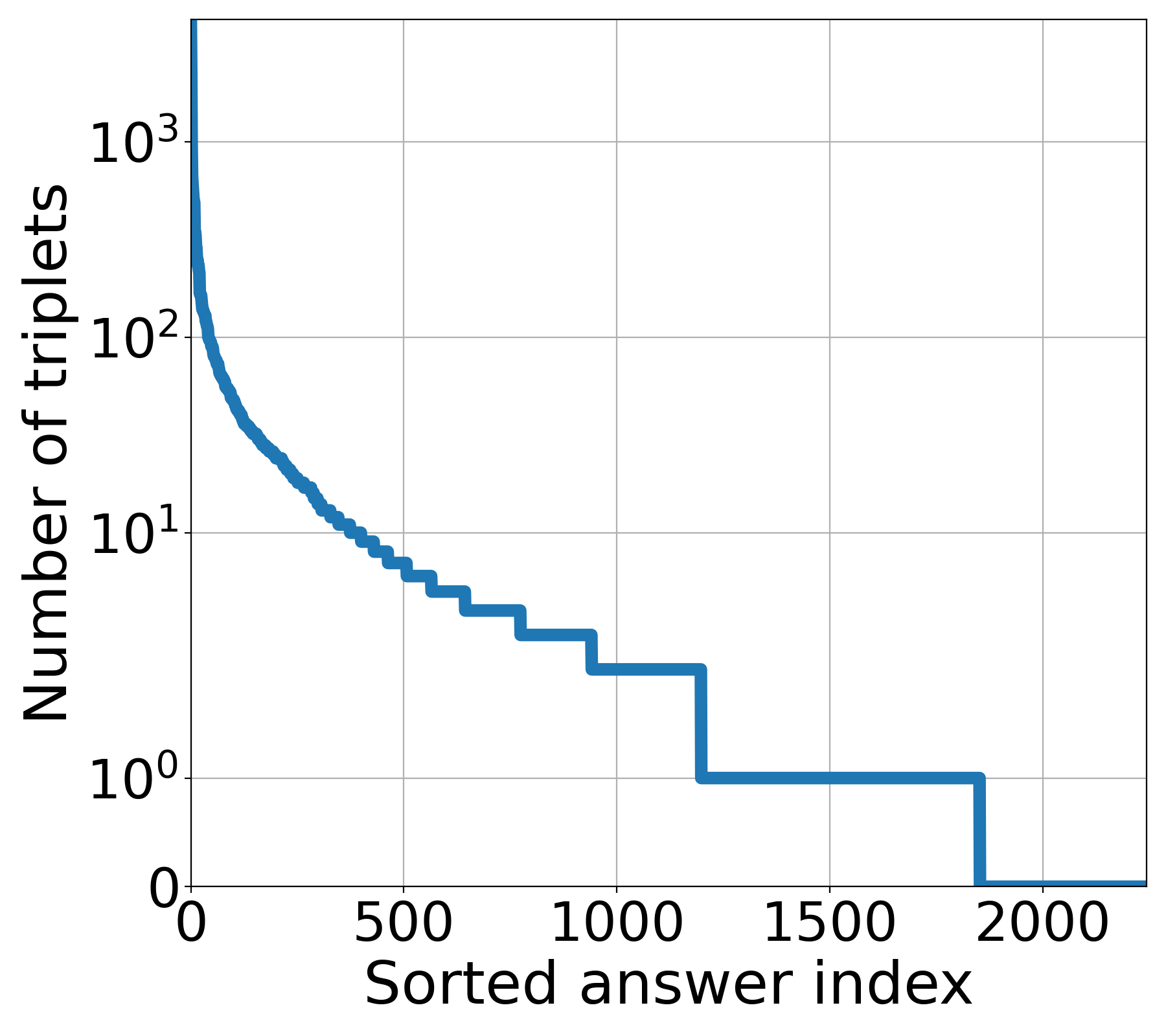}
        \caption{MSVD-QA}
        \label{fig:msvd_tail}
    \end{subfigure}
    \begin{subfigure}[t]{0.45\linewidth}
        \includegraphics[width=1.0\linewidth]{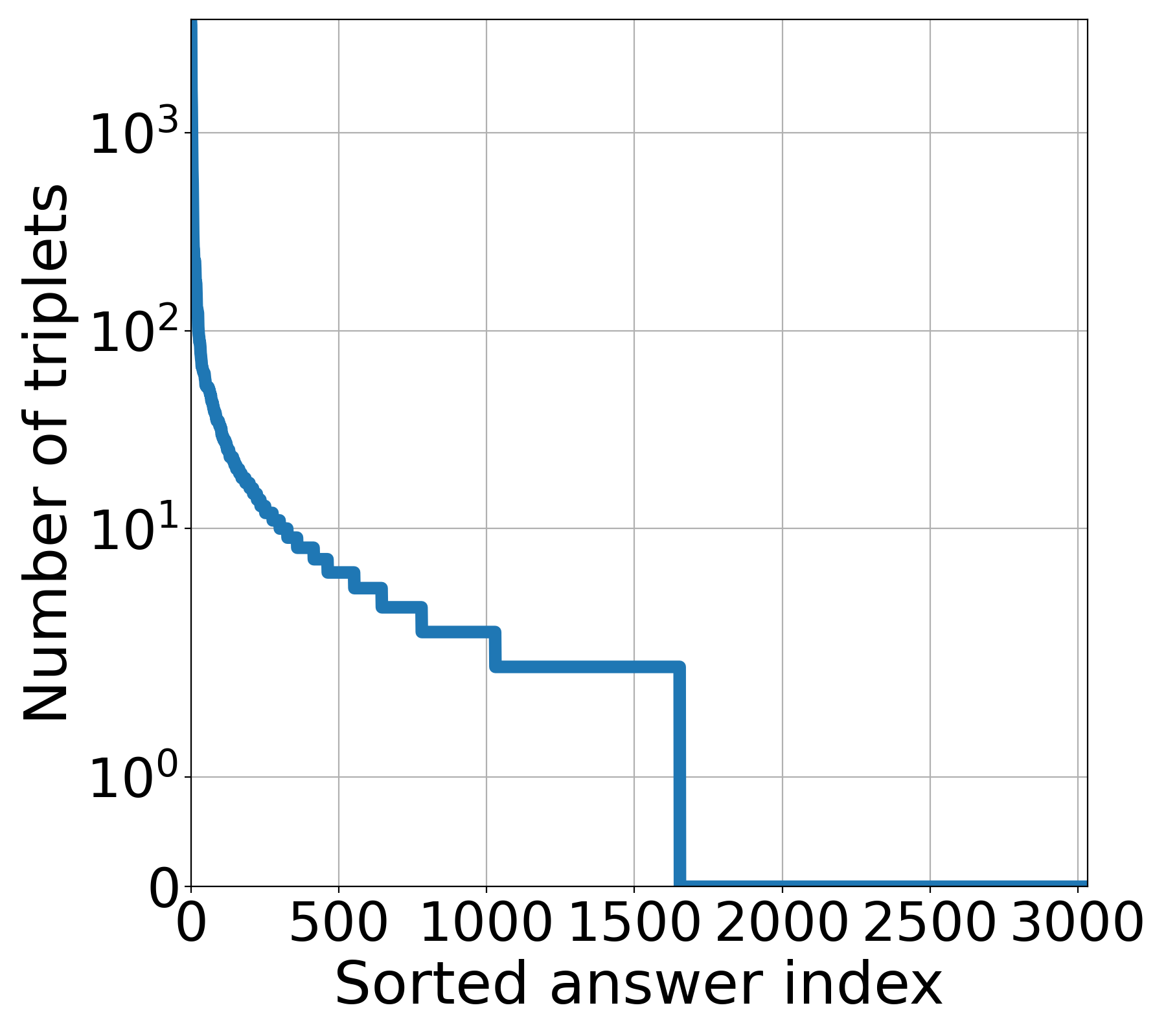}
        \caption{ActivityNet-QA}
        
        \label{fig:activitynet_tail}
    \end{subfigure}
        \begin{subfigure}[t]{0.45\linewidth}
        \includegraphics[width=1.0\linewidth]{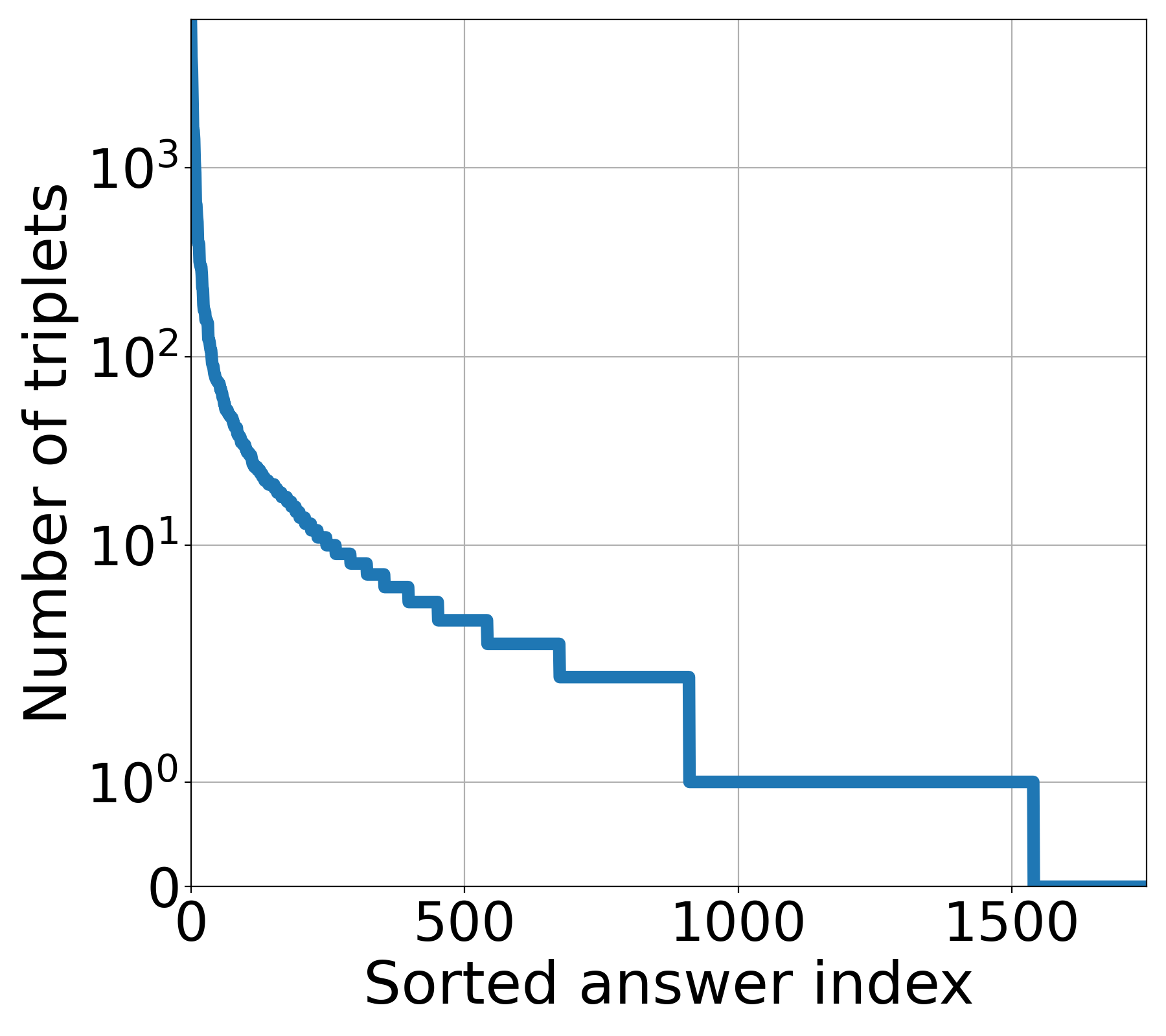}
        \caption{TGIF-QA}
        \label{fig:tgif_tail}
    \end{subfigure}
    \begin{subfigure}[t]{0.45\linewidth}
        \includegraphics[width=1.0\linewidth]{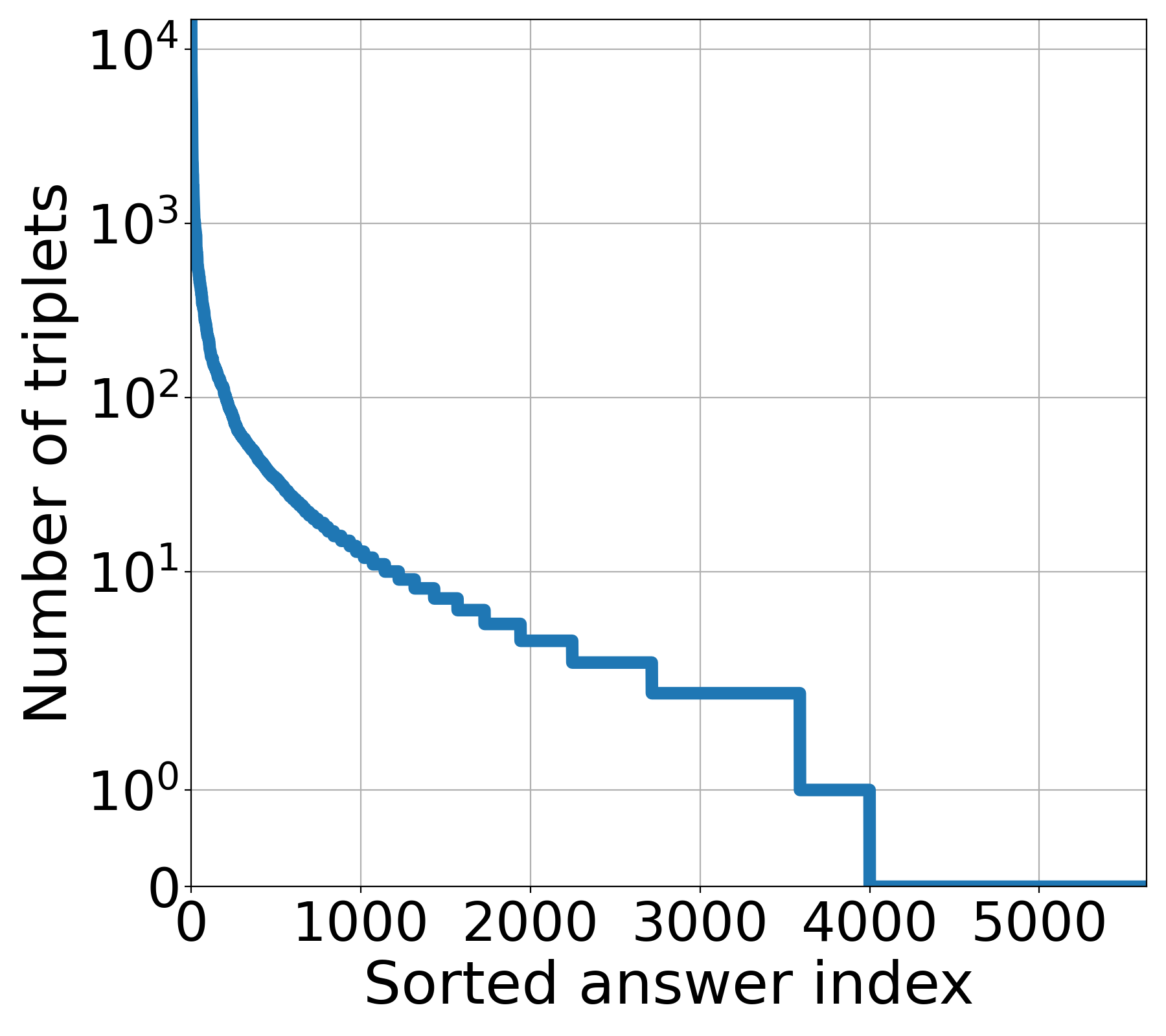}
        \caption{MSRVTT-QA}
        
        \label{fig:msrvtt_tail}
    \end{subfigure}
    \caption{\textbf{Dataset Statistics.}
    Sorted frequency statistics for each answer candidate reveal long tail distribution for all datasets.
    }
    \label{fig:longtail}
\end{figure}

Fig.~\ref{fig:venn_diagram} presents the distribution of answer candidates for the base, common, rare, and unseen answer categories in MSVD-QA, ActivityNet-QA, TGIF-QA, and MSRVTT-QA respectively. 
Note that the test answer candidates are composed mostly of rare and unseen answers, \textit{e.g.}, the number of rare and unseen answers (488 + 206) possess about 74\% of the test answer candidates (933) in TGIF.
In terms of base and common answers, most of them also appear in the test set.
Yet interestingly, for each dataset, more than half of the rare answers do not appear in the test set.
Furthermore, as depicted in Fig.~\ref{fig:longtail}, four datasets exhibit a long-tail answer distribution. 
Therefore, due to such imbalanced distribution, it is necessary to design the model under the open-vocabulary setting instead of the closed-vocabulary.
\begin{table*}[t!]
    \centering
    \setlength{\tabcolsep}{3.5pt}
    \begin{adjustbox}{width=\linewidth}
    \begin{tabular}{c| c c c c c c | c c c c c c | c c c c c c | c c c c c c }
        \toprule
        \textbf{Models} & \multicolumn{6}{c|}{\textbf{MSVD-QA}} & \multicolumn{6}{c|}{\textbf{ActivityNet-QA}} & \multicolumn{6}{c|}{\textbf{TGIF-QA}} & \multicolumn{6}{c}{\textbf{MSRVTT-QA}} \\

        & \textbf{B} & \textbf{C} & \textbf{R} & \textbf{U} & \textbf{T} & \cellcolor[HTML]{C0C0C0}\textbf{M} & \textbf{B} & \textbf{C} & \textbf{R} & \textbf{U} & \textbf{T} & \cellcolor[HTML]{C0C0C0}\textbf{M} & \textbf{B} & \textbf{C} & \textbf{R} & \textbf{U} & \textbf{T} & \cellcolor[HTML]{C0C0C0}\textbf{M} & \textbf{B} & \textbf{C} & \textbf{R} & \textbf{U} & \textbf{T} & \cellcolor[HTML]{C0C0C0}\textbf{M} \\
        \midrule
        \midrule
        \rowcolor[HTML]{FFF9C0}
        \multicolumn{25}{l}{\textbf{\textit{CVQA}}} \\
        Random & - & - & - & - & 0.1 & - & - & - & - & - & 0.1 & - & - & - & - & - & 0.1 & - & - & - & - & - & 0.1 & - \\ 
        CLIP~\cite{radford2021learning} & - & - & - & - & 7.2 & - & - & - & - & - & 1.2 & - & - & - & - & - & 3.6 & - & - & - & - & - & 2.1 & - \\
        JustAsk~\cite{yang2021just} & 17.1 & 10.1 & 12.8 & 0.0 & 13.5 & 7.0 & 19.9 & 8.6 & 8.3 & 0.0 & 12.3 & 2.8 & 28.4 & 10.4 & 9.9 & 0.0 & 23.8 & 6.9 & 5.9 & 5.5 & 5.5 & 0.0 & 5.6 & 3.3 \\
        FrozenBiLM~\cite{yang2022zero} & \textbf{46.4} & \textbf{26.6} & 12.6 & 0.0 & 33.7 & 9.9 & 44.1 & \textbf{17.9} & 7.4 & 0.0 & 25.9 & 3.8 & 48.9 & 27.4 & 11.0 & 0.0 & 41.9 & 11.5 & \textbf{19.3} & \textbf{13.9} & 0.0 & 0.0 & \textbf{16.7} & 3.2 \\
        \midrule
        \rowcolor[HTML]{FFF9C0}
        \multicolumn{25}{l}{\textbf{\textit{OVQA}}} \\
        \textbf{JustAsk+} & \cellcolor[HTML]{BFF2FF}{18.2} & \cellcolor[HTML]{BFF2FF}{12.9} & \cellcolor[HTML]{BFF2FF}{13.5} & \cellcolor[HTML]{BFF2FF}{13.1} & \cellcolor[HTML]{BFF2FF}{15.7} & \cellcolor[HTML]{BFF2FF}11.4 & \cellcolor[HTML]{FFD7D1}{12.8} & \cellcolor[HTML]{FFD7D1}{5.9} & \cellcolor[HTML]{FFD7D1}{6.2} & \textbf{\cellcolor[HTML]{BFF2FF}{6.7}} & \cellcolor[HTML]{FFD7D1}{9.4} & \textbf{\cellcolor[HTML]{BFF2FF}6.3} & \cellcolor[HTML]{BFF2FF}{29.5} & \cellcolor[HTML]{BFF2FF}{12.3} & \cellcolor[HTML]{BFF2FF}{12.7} & \textbf{\cellcolor[HTML]{BFF2FF}{13.2}} & \cellcolor[HTML]{BFF2FF}{25.3} & \cellcolor[HTML]{BFF2FF}11.9 & \cellcolor[HTML]{BFF2FF}{6.0} & \cellcolor[HTML]{FFD7D1}{5.2} & 5.5 & \textbf{\cellcolor[HTML]{BFF2FF}{4.6}} & \cellcolor[HTML]{BFF2FF}{5.8} & \cellcolor[HTML]{BFF2FF}4.5 \\
        \textbf{FrozenBiLM+} & \cellcolor[HTML]{FFD7D1}{46.3} & \textbf{26.6} & \textbf{\cellcolor[HTML]{BFF2FF}{16.5}} & \textbf{\cellcolor[HTML]{BFF2FF}{13.2}} & \textbf{\cellcolor[HTML]{BFF2FF}{34.9}} & \textbf{\cellcolor[HTML]{BFF2FF}13.7} & \textbf{\cellcolor[HTML]{BFF2FF}{45.3}} & \cellcolor[HTML]{FFD7D1}{17.3} & \textbf{\cellcolor[HTML]{BFF2FF}{8.9}} & \cellcolor[HTML]{BFF2FF}{3.1} & \textbf{\cellcolor[HTML]{BFF2FF}{27.3}} & \cellcolor[HTML]{BFF2FF}6.0 & \textbf{\cellcolor[HTML]{BFF2FF}{49.1}} & \textbf{\cellcolor[HTML]{BFF2FF}{27.6}} & \textbf{\cellcolor[HTML]{BFF2FF}{14.7}} & \cellcolor[HTML]{BFF2FF}{8.1} & \textbf{\cellcolor[HTML]{BFF2FF}{42.5}} & \textbf{\cellcolor[HTML]{BFF2FF}15.4} & \cellcolor[HTML]{FFD7D1}{15.5} & \cellcolor[HTML]{FFD7D1}{11.7} & \textbf{\cellcolor[HTML]{BFF2FF}{9.3}} & \cellcolor[HTML]{BFF2FF}{4.3} & \cellcolor[HTML]{FFD7D1}{14.1} & \textbf{\cellcolor[HTML]{BFF2FF}6.0} \\
        \bottomrule
    \end{tabular}
    \end{adjustbox}
    \caption{\textbf{Comparison with zero-shot state-of-the-art models.}
    }
    \label{tab:zero}
\end{table*}
\begin{table*}[t!]
    \centering
    \setlength{\tabcolsep}{3.5pt}
    \begin{adjustbox}{width=\linewidth}
    \begin{tabular}{c|c|c c c c c c | c c c c c c | c c c c c c | c c c c c c }
        \toprule
        \textbf{Models} & \multirow{2}{*}{\makecell{\textbf{Answer} \\ \textbf{encoder}}} & \multicolumn{6}{c|}{\textbf{MSVD-QA}} & \multicolumn{6}{c|}{\textbf{ActivityNet-QA}} & \multicolumn{6}{c|}{\textbf{TGIF-QA}} & \multicolumn{6}{c}{\textbf{MSRVTT-QA}} \\

        & & \textbf{B} & \textbf{C} & \textbf{R} & \textbf{U} & \textbf{T} & \cellcolor[HTML]{C0C0C0}\textbf{M} & \textbf{B} & \textbf{C} & \textbf{R} & \textbf{U} & \textbf{T} & \cellcolor[HTML]{C0C0C0}\textbf{M} & \textbf{B} & \textbf{C} & \textbf{R} & \textbf{U} & \textbf{T} & \cellcolor[HTML]{C0C0C0}\textbf{M} & \textbf{B} & \textbf{C} & \textbf{R} & \textbf{U} & \textbf{T} & \cellcolor[HTML]{C0C0C0}\textbf{M} \\
        \midrule
        \midrule
        
        \multirow{2}{*}{\textbf{All-in-one+}} & CLIP & 62.4 & 24.3 & 0.5 & 0.1 & 40.1 & 5.3 & 64.4 & 25.9 & 0.6 & 0.2 & 36.7 & 2.6 & 77.3 & 29.7 & 2.0 & 0.0 & 63.0 & 8.0 & 49.3 & 7.8 & 0.2 & \textbf{0.0} & 37.9 & 2.8 \\
        & DeBERTa & \textbf{62.8} & \textbf{34.0} & \textbf{6.3} & \textbf{0.4} & \textbf{43.8} & \textbf{9.4} & \textbf{64.9} & \textbf{35.9} & \textbf{9.8} & \textbf{0.5} & \textbf{40.2} & \textbf{6.8} & \textbf{78.3} & \textbf{39.3} & \textbf{10.2} & \textbf{0.4} & \textbf{66.0} & \textbf{13.2} & \textbf{49.8} & \textbf{14.6} & \textbf{1.6} & \textbf{0.0} & \textbf{39.5} & \textbf{4.7} \\

        \midrule
        \multirow{2}{*}{\textbf{VIOLET+}} & CLIP & {68.0} & {31.0} & {1.5} & \textbf{0.1} & {45.5} & {7.4} & \textbf{64.3} & {33.8} & {2.6} & {0.1} & {38.6} & {3.9} & {76.3} & {29.4} & {2.5} & {0.0} &  {62.4} & {8.8} & {52.7} & {7.4} & {0.4} & \textbf{0.0} & {40.3} & {3.0} \\
        & DeBERTa & \textbf{70.6} & \textbf{38.8} & \textbf{6.7} & \textbf{0.1} & \textbf{49.5} & \textbf{10.7} & {63.4} & \textbf{37.1} & \textbf{9.2} & \textbf{0.6} & \textbf{39.7} & \textbf{6.1} & \textbf{77.3} & \textbf{38.9} & \textbf{10.8} & \textbf{2.0} & \textbf{65.3} & \textbf{14.3} & \textbf{53.8} & \textbf{14.7} & \textbf{0.9} & \textbf{0.0} & \textbf{42.4} & \textbf{4.5} \\

        \bottomrule
    \end{tabular}
    \end{adjustbox}
    \caption{\textbf{Ablation study on the answer encoder type.}
    }
    \label{tab:encoder}
\end{table*}
\section{Implementation details}

\noindent \textbf{All-in-one~\cite{wang2022all}.}
The model is fine-tuned on four datasets with a batch size of 512 for 20 epochs. 
The learning rate is 1e-4 with a warm up step of 10\% of the total iterations.
AdamW optimizer~\cite{loshchilov2017decoupled} is used.
For video features, 3 video frames are randomly sampled and resized to $224 \times 224$. 
Then each frame is split into patches of size $14 \times 14$. 
In the setting of CVQA, the number of training and test answers are identical to one another with MSVD 1000, MSRVTT is 1500, ActivityNet is 1000, and TGIF is 1540.

\noindent \textbf{VIOLET~\cite{fu2021violet}.}
 For all experiments, we employ the AdamW with $\beta=(0.9, 0.98)$, and the initial learning rate is set to 1.2e-5. 
 The weight decay is 1e-3. 
 The number of video frames sampled is 5 with the size of $224 \times 224$ and are split into patch sizes of $32 \times 32$. 
 The batch size used for MSVD, MSRVTT, TGIF, and ActivityNet is 10, 12, 10, and 8 per GPU respectively. 
 For training the model in CVQA, the number of answers used for testing and training is consistent with MSVD 1000, MSRVTT 1500, TGIF 1540, and ActivityNet 1654.
 
\noindent \textbf{JustAsk~\cite{yang2021just}.}
Fine-tuning for the model is implemented for 20 epochs and we use Adam~\cite{kingma2014adam} optimizer with a batch size of 256 and validation batch size of 2048. 
For the learning rate, we utilize the cosine annealing scheduler with an initial value of 1e-5. 
The video features are equally space sampled and padded up to a maximum of 20. 
The dimension of the video feature is 1024, the text is 768 and the final embedding is 512. 
The Dropout~\cite{srivastava2014dropout} probability is set to 0.1. 
The number of training and test answers for CVQA is MSVD 1852, MSRVTT 4000, TGIF 1540, and ActivityNet 1654. 

\noindent \textbf{FrozenBiLM~\cite{yang2022zero}.}
For each video and text encder, we use $T = 10$ for the number of frames and $N = 256$ for the number of text tokens.
Each frame is resized to the size of $224 \times 224$ and its feature is extracted by CLIP ViT-L/14~\cite{radford2021learning,dosovitskiy2020image}.
We use a hidden dimension size of $D = 1536$.
Learning rate is set to 5e-5 and linear warm up is applied for the first 10\% of total iterations.
After the warm up, a learning rate is decayed to 0 for the remaining iterations.
We train the model with a batch size of 32 during 20 epochs for all the datasets.
Dropout probability is 0.1 and Adam optimizer of $\beta=(0.9, 0.95)$ is adapted with no weight decay.
\section{Additional quantitative results}

\subsection{Zero-shot performance}
We compare the zero-shot performances between the standard CVQA baselines and our developed OVQA baselines in Tab.~\ref{tab:zero}.
On MSVD, ActivityNet and TGIF, our FrozenBiLM+ outperforms the standard FrozenBiLM by 1.2\%, 1.4\%, and 0.6\% on the total performance (\textbf{T}), achieving state-of-the-art results.
Also for all the datasets, mAcc (\textbf{M}) on both JustAsk+ and FrozenBiLM+ are improved by a large margin.
This implies that considering rare and unseen answers by fully leveraging the generalizability of backbone models pretrained on the large-scale dataset also improves the zero-shot performance.

\subsection{Ablation studies}

\noindent \textbf{Answer encoder type.}
We conduct an ablation study on the answer encoder type by comparing CLIP~\cite{radford2021learning} and DeBERTa~\cite{he2020deberta} in Tab.~\ref{tab:encoder}.
In general, adopting DeBERTa outperforms CLIP by a large margin especially on mAcc (\textbf{M}) for all datasets.

\noindent \textbf{Effectiveness of $\boldsymbol{\varepsilon}$.}
In Tab.~\ref{tab:epsilon}, we also experiment by adjusting the $\varepsilon$ in Eq. (7) of the main paper on FrozenBiLM+.
Note that with a wide range of $\varepsilon \in [0.3, 0.9]$, our method equipped with the GNN-based soft verbalizer shows superior performance to the standard FronzeBiLM ($\varepsilon = 1.0$).
\begin{table}[t!]
    \centering
    \setlength{\tabcolsep}{3.5pt}
    \begin{tabular}{c|c c c c c >{\columncolor{lightgray}}c}
        \toprule
        $\boldsymbol{\varepsilon}$ & \multicolumn{6}{c}{\textbf{ActivityNet}} \\
        & \textbf{B} & \textbf{C} & \textbf{R} & \textbf{U} & \textbf{T} & \textbf{M} \\
        \midrule
        \midrule
        1.0 & 67.7 & 37.4 & 15.5 & 4.2 & 43.2 & 10.4 \\
        0.9 & \textbf{68.7} & 37.3 & 15.2 & 4.5 & 43.7 & 10.7 \\
        0.8 & 67.8 & 38.6 & 16.9 & 4.7 & 43.8 & 11.1 \\
        0.7 & 68.2 & \textbf{39.9} & \textbf{18.5} & \textbf{5.8} & \textbf{44.6} & \textbf{11.9} \\
        0.6 & 68.1 & 38.7 & 17.6 & 5.1 & 44.1 & 11.7 \\
        0.5 & 67.5 & 38.4 & 16.2 & 4.9 & 43.6 & 11.1 \\
        0.4 & 68.3 & 37.8 & 15.6 & 5.3 & 43.8 & 11.1 \\
        0.3 & 68.2 & 36.8 & 14.9 & 5.2 & 43.4 & 11.2 \\
        0.2 & 68.2 & 36.3 & 13.1 & 5.1 & 43.1 & 10.3 \\
        0.1 & 68.3 & 35.5 & 12.5 & 4.1 & 42.7 & 9.3 \\
        0.0 & 66.2 & 34.9 & 12.2 & 4.2 & 41.6 & 9.3 \\
        \bottomrule
    \end{tabular}
    \caption{\textbf{Ablation study on $\varepsilon$.}
    }
    \label{tab:epsilon}
\end{table}
\begin{figure*}[t] 
    \centering
    \begin{subfigure}[t]{0.49\linewidth}
        \includegraphics[width=1.0\linewidth]{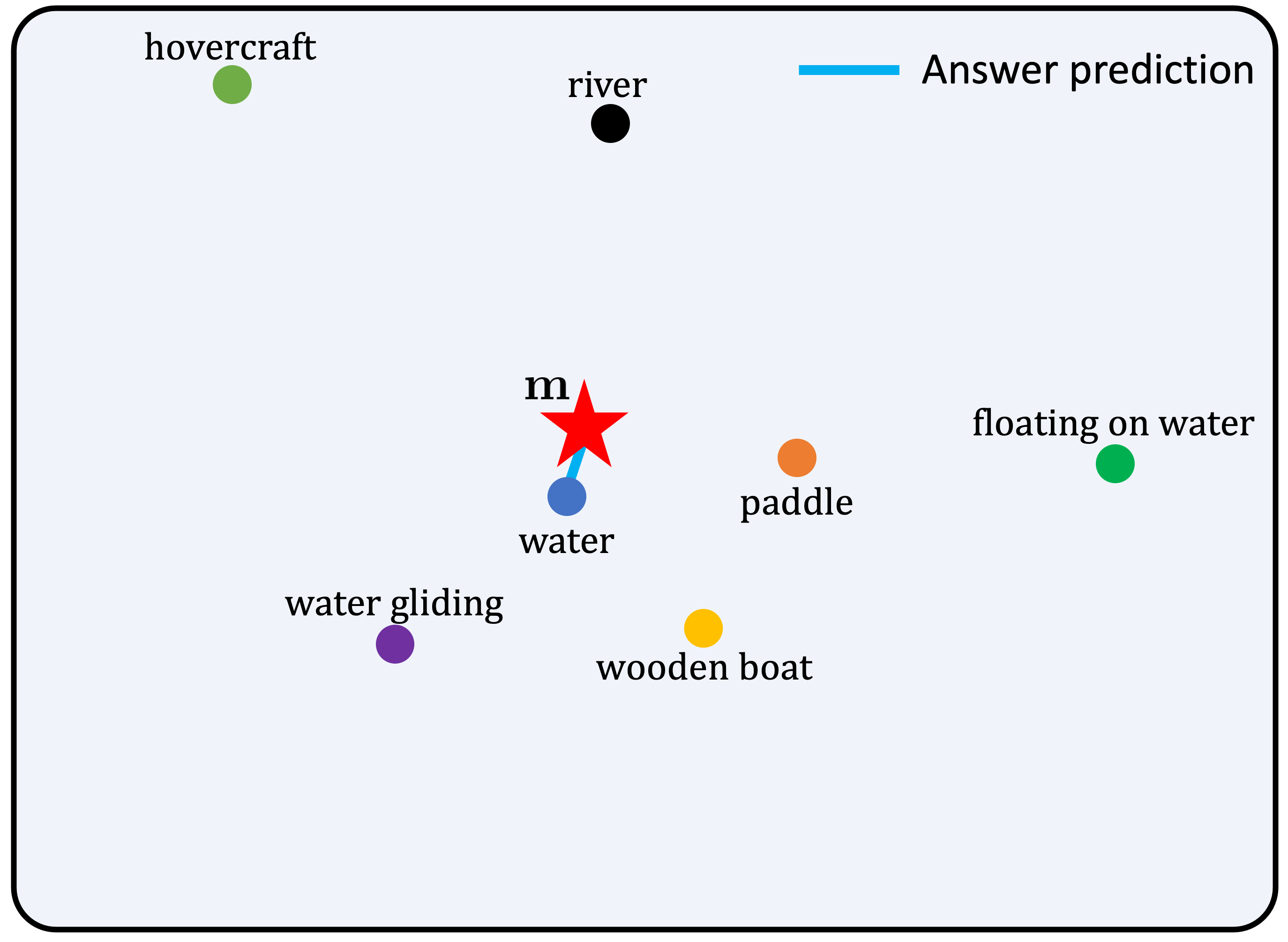}
        \caption{FrozenBiLM+ without GNN-based soft verbalizer}
        \label{fig:before}
    \end{subfigure}
    \begin{subfigure}[t]{0.49\linewidth}
        \includegraphics[width=1.0\linewidth]{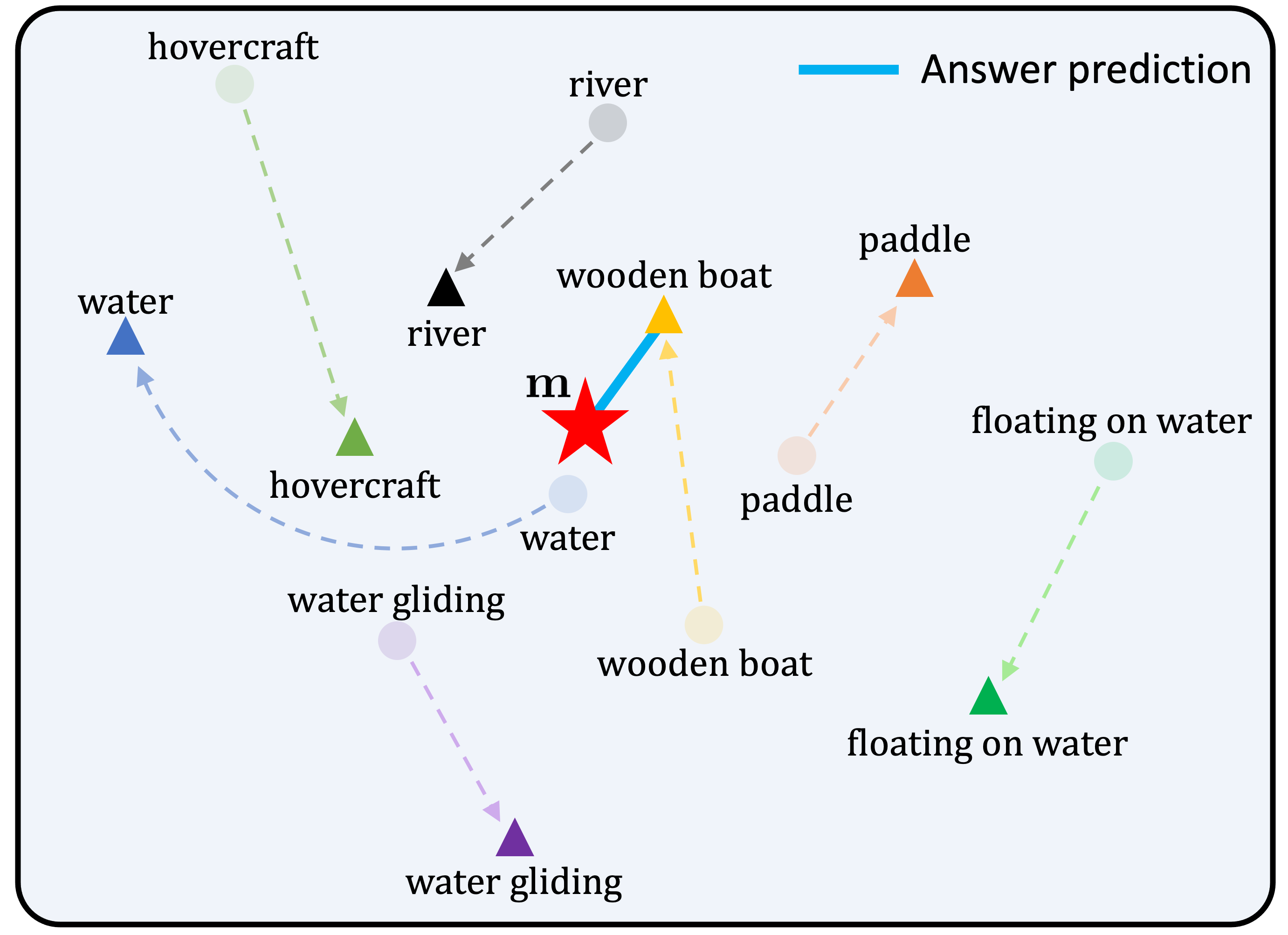}
        \caption{FrozenBiLM+ with GNN-based soft verbalizer}
        \label{fig:after}
    \end{subfigure}
    \caption{\textbf{TSNE of answer embeddings before/after adapting GNN-based soft verbalizer.}
    $\mathbf{m}$ is an output feature of the [MASK] token.
    The prediction of the model is changed from ``water'' in (a) to ``wooden boat'' in (b).
    }
    \label{fig:embeddings}
\end{figure*}
\begin{figure}[t!] 
    \centering
    \begin{subfigure}[t]{0.44\linewidth}
        \includegraphics[width=1.0\linewidth]{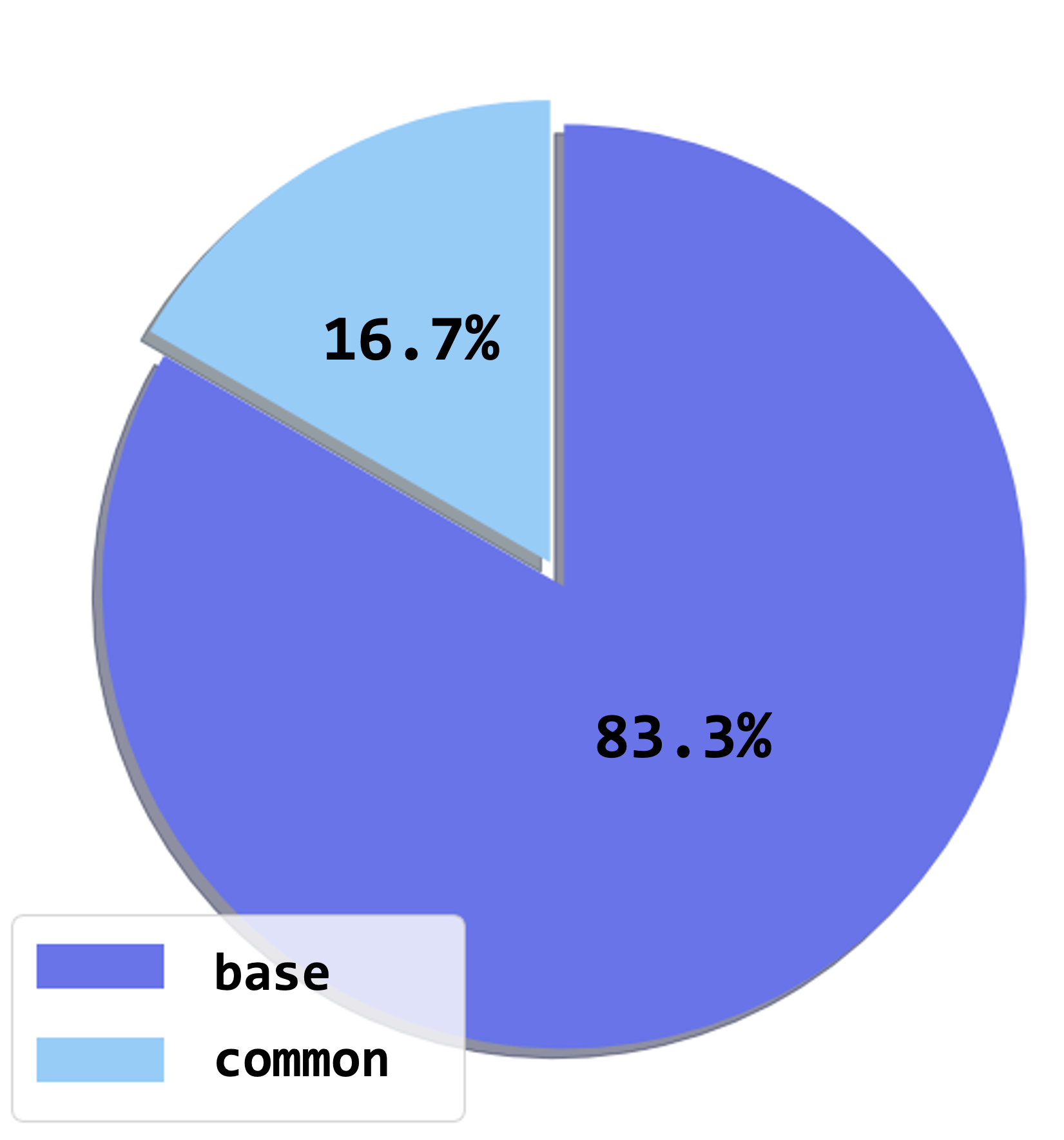}
        \caption{VIOLET}
        \label{fig:pie_base}
    \end{subfigure}
    \begin{subfigure}[t]{0.45\linewidth}
        \includegraphics[width=1.0\linewidth]{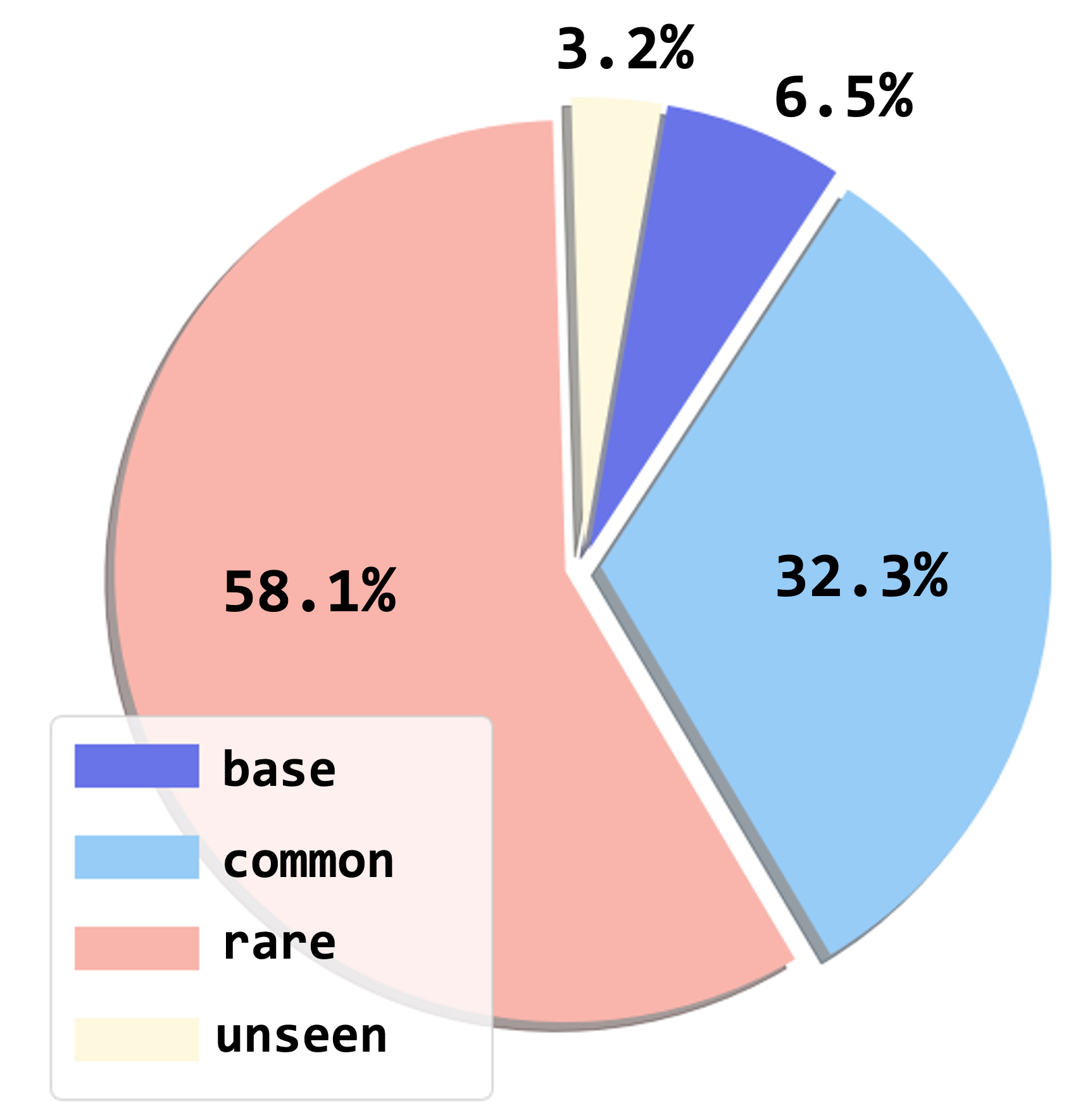}
        \caption{VIOLET+}
        \label{fig:pie_ours}
    \end{subfigure}
    \caption{\textbf{Proportion of answer categories with an accuracy of 90\%.}
    The portion of answer categories in TGIF that (a) VIOLET and (b) VIOLET+ achieve an accuracy of 90\%. 
    }
    \label{fig:pie}
\end{figure}
\section{Additional qualitative results}

\subsection{Comparison of answer category proportion}
We analyze the answers that VIOLET and VIOLET+ correctly predict. 
Fig.~\ref{fig:pie} shows the proportion of answer categories that are predicted by VIOLET and VIOLET+ with an accuracy of 90\% or higher.
VIOLET in Fig.~\ref{fig:pie_base} focuses on base and common categories, and the portion of the base category answers is 83.3\%. 
On the other hand, Fig.~\ref{fig:pie_ours} shows that VIOLET+ accurately predicts the answers in the rare and unseen categories beyond base and common answers.
The portion of rare and unseen categories significantly increased. 
This evidences that the bias of VIOLET toward frequent answers is alleviated in VIOLET+.

\subsection{Answer embeddings visualization}
\begin{figure}[t] 
    \centering
        \includegraphics[width=0.99\linewidth]{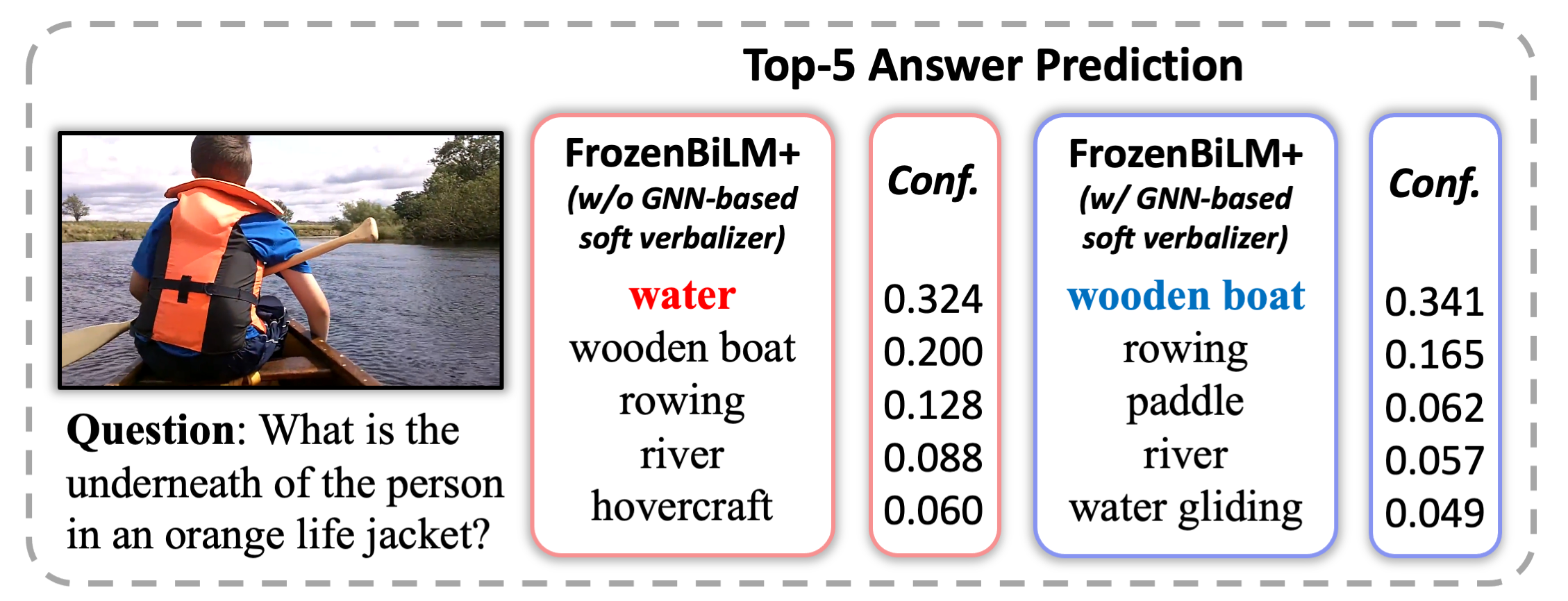}
        \caption{\textbf{Confidence scores of the top-5 predictions w/ and w/o GNN-based soft verbalizer on FrozenBiLM+.}}
        \label{fig:apdx_top5}
\end{figure}

Fig.~\ref{fig:apdx_top5} illustrates another qualitative example of the model with and without a GNN-based soft verbalizer on FrozenBiLM+.
GNN-based soft verbalizer successfully corrects the prediction from ``water'' to ``wooden boat''.
Also, in Fig.~\ref{fig:embeddings}, we visualize TSNE of answer embedding changes before/after adapting GNN-based soft verbalizer in the above example.
Fig.~\ref{fig:before} shows that the model predicts ``water'', which is the closest answer to $\mathbf{m}$, as the answer without a GNN-based soft verbalizer.
On the other hand, in Fig.~\ref{fig:after}, GNN-based soft verbalizer effectively updates the answer embeddings by moving the embedding of ``wooden boat'' close to $\mathbf{m}$, and the prediction is corrected to ``wooden boat''.

\begin{table}[t!]
    \centering
    \setlength{\tabcolsep}{3.5pt}
    \begin{adjustbox}{width=\linewidth}
    \begin{tabular}{c|c c|c c|c c|c c}
        \toprule
        \textbf{Models} & \multicolumn{2}{c|}{\textbf{MSVD}} & \multicolumn{2}{c|}{\textbf{ActivityNet}} & \multicolumn{2}{c|}{\textbf{TGIF}} & \multicolumn{2}{c}{\textbf{MSRVTT}} \\
        & \textbf{BNG}$\downarrow$ & \textbf{M}$\uparrow$ & \textbf{BNG}$\downarrow$ & \textbf{M}$\uparrow$ & \textbf{BNG}$\downarrow$ & \textbf{M}$\uparrow$ & \textbf{BNG}$\downarrow$ & \textbf{M}$\uparrow$ \\
        \midrule
        \midrule
        All-in-one~\cite{wang2022all} & 41.3 & 7.9 & 49.1 & 5.3 & 56.0 & 10.1 & 42.2 & 3.9 \\
        \textbf{All-in-one+} & \textbf{39.3} & \textbf{9.4} & \textbf{47.3} & \textbf{6.8} & \textbf{50.6} & \textbf{13.2} & \textbf{39.9} & \textbf{4.7} \\
        \midrule
        VIOLET~\cite{fu2021violet} & 70.7 & 2.7 & 49.6 & 3.7 & 77.9 & 4.5 & 54.6 & 1.4 \\
        \textbf{VIOLET+} & \textbf{44.2} & \textbf{10.7} & \textbf{46.1} & \textbf{6.1} &  \textbf{49.2} & \textbf{14.3} & \textbf{43.9} & \textbf{4.5} \\
        \midrule
        JustAsk~\cite{yang2021just} & 38.5 & 12.6 & 41.2 & 8.2 & 44.9 & 11.7 & 38.2 & 7.0 \\
        \textbf{JustAsk+} & \textbf{37.2} & \textbf{14.5} & \textbf{39.5} & \textbf{11.5} & \textbf{43.5} & \textbf{14.4} & \textbf{37.8} & \textbf{7.6} \\
        \midrule
        FrozenBiLM~\cite{yang2022zero} & 37.4 & 17.2 & 47.3 & 7.9 & 37.8 & 23.5 & 40.2 & 6.7 \\
        \textbf{FrozenBiLM+} & \textbf{35.0} & \textbf{21.3} & \textbf{46.6} & \textbf{11.9} & \textbf{35.0} & \textbf{30.2} & \textbf{35.7} & \textbf{12.2} \\
        \bottomrule
    \end{tabular}
    \end{adjustbox}
    \caption{\textbf{Comparison of Base and Non-base performance gap (BNG).}
    }
    \label{tab:bng}
\end{table}
\section{A new metric to measure the model bias}

We here introduce a new metric, Base and Non-base performance Gap (BNG).
BNG evaluates how much the model is biased toward base answers, and is calculated as:
\begin{equation}
    \text{BNG (\%)} = \text{Base (\%)} - \text{Non-base (\%)},
\end{equation}
where Non-base consists of common, rare, and unseen answers.
The lower BNG indicates that the model has less bias.
In Tab.~\ref{tab:bng}, our developed baselines outperforms previous CVQA baselines by a large margin in terms of BNG as well as mAcc (\textbf{M}).
Especially, by comparing VIOLET and VIOLET+, the BNG is decreased by 26.5\% and 28.7\% on MSVD and TGIF respectively, and mAcc (\textbf{M}) is also improved by 8\% and 9.8\%.
This implies that the model bias toward frequent answers is effectively alleviated on VIOLET+.

\end{document}